\documentclass[conference]{IEEEtran}

\newcommand{\subparagraph}{}
\usepackage{titlesec}
\titlespacing*{\section}
{0pt}{1ex plus 1ex minus .2ex}{1ex plus .2ex}
\titlespacing*{\subsection}
{0pt}{0.5ex plus 1ex minus .2ex}{0.5ex plus .2ex}

\IEEEoverridecommandlockouts
\usepackage{booktabs}
\usepackage{subfig}
\usepackage{graphicx}
\usepackage{booktabs}
\usepackage{epstopdf}
\usepackage{multirow}
\usepackage{array}
\usepackage{enumitem}

\usepackage{tcolorbox}

\usepackage{mathtools}

\usepackage{algorithmic,algorithm}

\usepackage{url}
\usepackage{bm}
\usepackage{amsbsy}
\usepackage{amsmath}
\usepackage{amssymb}
\usepackage{amsthm}
\usepackage{amsfonts}
\usepackage{amstext}
\usepackage{bbm}

\DeclarePairedDelimiterX{\infdivx}[2]{(}{)}{%
  #1\;\delimsize\|\;#2%
}
\newcommand{\kldiv}{\mbox{KL-DIV}\infdivx}

\begin{document}
\title{Securing Behavior-based Opinion Spam Detection}
\author{
\IEEEauthorblockN{Shuaijun Ge\IEEEauthorrefmark{1},
Guixiang Ma\IEEEauthorrefmark{2},
Sihong Xie\IEEEauthorrefmark{3} and
Philip S. Yu\IEEEauthorrefmark{4}}
\IEEEauthorblockA{
\IEEEauthorrefmark{1}dorisgebupt@gmail.com,
\IEEEauthorrefmark{2}gma4@uic.edu,
\IEEEauthorrefmark{3}six316@lehigh.edu,
\IEEEauthorrefmark{4}psyu@uic.edu}

    \\[-4.0ex]
}

\maketitle

\begin{abstract}
Reviews spams are prevalent in e-commerce to manipulate product ranking and customers decisions maliciously.
While spams generated based on simple spamming strategy can be detected effectively,
hardened spammers can evade regular detectors via more advanced spamming strategies.
Previous work gave more attention to evasion against text and graph-based detectors,
but evasions against behavior-based detectors are largely ignored,
leading to vulnerabilities in spam detection systems.
Since real evasion data are scarce,
we first propose EMERAL
(Evasion via Maximum Entropy and Rating sAmpLing)
to generate evasive spams to certain existing detectors.
EMERAL can simulate spammers with different goals and levels of knowledge about the detectors,
targeting at different stages of the life cycle of target products.
We show that, in the evasion-defense dynamic,
only a few evasion types are meaningful to the spammers,
and any spammer will not be able to evade too many detection signals at the same time.
We reveal that some evasions are quite insidious and can fail all detection signals.
We then propose DETER (Defense via Evasion generaTion using EmeRal),
based on model re-training on diverse evasive samples generated by EMERAL.
Experiments confirm that DETER 
is more accurate in detecting both suspicious time window 
and individual spamming reviews.
In terms of security, DETER is versatile enough to
be vaccinated against diverse and unexpected evasions,
is agnostic about evasion strategy and can be released without privacy concern.
\end{abstract}


\section{Introduction}
\label{sec:intro}
Many customers post reviews on online commerce websites such as Amazon and Yelp.
the opinionated reviews
help shape product ranking and reputation,
find consumers high-quality products,
and make products become more visible via word-of-mouth~\cite{duan2008, Dellarocas2005, Judith2006}.
However, such a mechanism has also attracted
many dishonest businesses to hire professional spammers
to post ungrounded reviews 
to manipulate product reputations~\cite{jindal2008, ftccharge_2010, bbcsamsung, webmdarticle}.
Customers can be misled to low quality products,
honest businesses can suffer from unfair competition,
and the whole online e-commerce can be rendered less trustworthy.

To combat opinion spams,
prior works have proposed abundant different detection models
based on texts~\cite{lim2010, Jindal2007, jindal2008, Xu2012},
user-behaviors~\cite{lim2010, Xie2012a, Fei13},
network structures~\cite{Akoglu2013, Wang2012, Lauw2006}.
However, more resourceful spammers can exploit information about the detectors available through publications, spam-spotting guidance
and detection websites\footnote{\url{https://www.fakespot.com/},
and \url{https://reviewmeta.com/}},
to craft insidious spamming campaigns that can evade
graph-based and text-based detectors~\cite{Chen2017, hooi2016fraudar}.
However, adversarial evasions against behavior-based detectors
have so far received less attention.
This leads to potential vulnerabilities in spam detection systems that integrate behavior-based detectors.


Addressing this gap is non-trival, however.
First, a deployed detectors can be subject to adversarial probing and attack.
For example, a spammer can gather knowledge about training data, features and models
of the detector and engineer evasive attacks against the probed detector~\cite{MukherjeeV0G13,Zhu2016}.
Also, diverse spamming strategies are likely to be adopted simultaneously by multiple spammers.
These scenarios lead to 
attack-defense strategy asymmetry --- the defending strategy
is not optimal with respect to the actual attacking strategy,
and a detector assuming a fixed evasion strategy~\cite{liang2017, Bruckner2009, Lowd2005} is more vulnerable.
Ideally, a detector has to be agnostic of any spamming strategies, but
the simple solution of blindly reacting to anomalies of any detection signals
can produce too many false positives (see the experiments).

Model retraining can obtain unseen but probable attacks
to hardened the detector against future attacks without assuming a single fixed attacking strategy.
The key is to generate spamming actions
to quantitatively manipulate detection signals under certain domain constraints.
In the spam detection application,
existing evasion attacks adopt
closed-form or differentiable objective functions~\cite{Lowd2005, Biggio2013, Nelson2010, hooi2016fraudar, Chen2017}.
In malware detection, regardless of the target detector,
direct manipulation and feature-sampling mapping were adopted,
with domain constraints preserved~\cite{Xu2016, liang2017, Maiorca2013} or totally ignored~\cite{Srndic2014}.
evasions against classification models~\cite{Biggio2013, Kantchelian2016} are usually Generated in the feature space without
constraints from the application domains.
Crafting real spamming attacks under constraints
is not pertaining to the high-level detection models and
is thus more fundamental and challenging.
For example, an attack can post all 5-star reviews to
boost the overall rating of a product from 2 to 4 stars,
but the attack will have a detectable skew rating distribution,
due to the constraint over changes in average rating and attack rating distribution.
While genetic algorithms~\cite{Xu2016, liang2017} can hypothetically modify previous spamming campaigns
for evasion, the approach is not scalable and requires a known spamming attack which is usually not available.

To address the challenges,
we first identify, as the target of the spammer, a set of
detection signals~\cite{Ye2016, Minnich2015, MukherjeeV0G13, Rayana2015Collective, Xie2012a, Fei13}
that characterize spammer behaviors.
We propose ``EMERAL'', a maximum entropy model
to quantitatively encode the spammers' knowledge, objective and domain constraints.
By solving the resulting optimization problem we obtain an optimal attacking vector
that further guide the generation of real evasive spams.
The model,
captures explicitly capture the quantitative dependencies among multiple detection signals for realistic
attack generation.
The model is general,
as multiple types of evasions against behavior-based detection signals 
can be included as objectives or constraints during different stage of the life-cycle
of a product.

With EMERAL,
we propose a novel defense, DETER, based on retraining,
where training data containing possible future evasive spams are first generated by EMERAL
and then used to train more effective detector without assuming a fixed evasion strategy.
Based on the weights learned by DETER and the properties of evasion generation,
DETER can be released to the spammer without security concern. 
Experimentally, the new defense is shown to be superior
to any fixed single detection signals, simple signal aggregation and even ensembles of multiple classifiers trained on the same adversarial examples.

\section{Detection and threat models}
\label{sec:prelim}

A review system has a set of accounts $\mathcal{U}=\{u_1,\dots,u_n\}$,
items ${\mathcal V}=\{v_1,\dots, v_m\}$,
and reviews ${\mathcal R}=\{r_{ij}: i\in \{1,\dots, n\}, j\in \{1,\dots, m\}\}$,
where $r_{ij}$ is the review posted by account $u_i$ to item $v_j$.
$r_{ij}$ contains
its text contents $c(r_{ij})$, its rating $s(r_{ij})$ and its posting time $t(r_{ij})$.
We focus on detection model based on aggregated rating behaviors
over time~\cite{Feng2012, Xie2012a, Fei13, Santosh2016, Ye2016}:
reviews in ${\mathcal R}$ are grouped into windows and
for each window, numeric detection signals in Table~\ref{tab:detection_signals}
are computed to obtain window suspicious scores.
These window-wise signals
are unique and not available in detection on the review, reviewer and item level, and
can help detect individual reviews~\cite{Feng2012}.
We focus on spammers, be it human or bots, with the goal
of promoting the target products' long and short term reputation, measured 
in cmulative average rating (CAR) and current month ranking (CMR, defined as the ranking of a business, among all businesses,
based on the current month's average rating~\cite{tripadvisor2013}).
CAR and CMR are shown to be vulnerable to spammers' manipulations~\cite{Lappas2016,Tucker2011,Ghose2013}.
The demoting spams can be handled similarly by the proposed models.
We first introduce behavior-based detection signals defined by previous work.

\begin{table}[t]
\centering
\small
\caption{Abnormal reviewer behavior detection signals.}
\label{tab:detection_signals}
\resizebox{0.8\columnwidth}{!}{
\begin{tabular}{ccc}
\toprule
\textbf{Signal names} & \textbf{Suspicious when} & \textbf{Descriptions}\\

\cmidrule[0.4pt](lr{0.125em}){1-1}
\cmidrule[0.4pt](lr{0.125em}){2-2}
\cmidrule[0.4pt](lr{0.125em}){3-3}

\multirow{2}{*}{NR ($\Delta$NR) }& \multirow{2}{*}{H (H)} & Number of reviews and change \\
	& & of NR in a window~\cite{Xie2012a}.\\

\cmidrule[0.4pt](lr{0.125em}){2-2}
\cmidrule[0.4pt](lr{0.125em}){3-3}

\multirow{1}{*}{$\Delta$CAR}& \multirow{1}{*}{H (H)} & Change in Cumulative Average Rating\\

\cmidrule[0.4pt](lr{0.125em}){2-2}
\cmidrule[0.4pt](lr{0.125em}){3-3}

\multirow{2}{*}{CAR-DEV} & \multirow{2}{*}{H} &Deviation of CAR from\\
						& &its predicted value~\cite{Ye2016}.\\

\cmidrule[0.4pt](lr{0.125em}){2-2}
\cmidrule[0.4pt](lr{0.125em}){3-3}

\multirow{2}{*}{NPR ($\Delta$NPR)} & \multirow{2}{*}{H} &Number of positive  \\
						& & reviews and its changes.\\

\cmidrule[0.4pt](lr{0.125em}){2-2}
\cmidrule[0.4pt](lr{0.125em}){3-3}

\multirow{2}{*}{EN ($\Delta$EN)} & \multirow{2}{*}{L (H)} & Entropy of ratings (and its  \\
 & & change) in each window.\\

\cmidrule[0.4pt](lr{0.125em}){2-2}
\cmidrule[0.4pt](lr{0.125em}){3-3}

\multirow{3}{*}{KL-DIV} & \multirow{3}{*}{H} & KL-divergence between rating\\
		& & distribution of a window\\
		& & and historic distribution.\\
\bottomrule

\end{tabular}
 }
\end{table}

\subsection{Time series based detection signals}
\label{sec:ar_model}
Normal review traffic shall arrive in a smooth manner while spamming reviews usually
arrive in a more abrupt pattern~\cite{Xie2012a, Fei13}.
Besides, to effectively promote product reputation,
spammers also aim at lifting
the average rating of the targets significantly~\cite{duan2008, Dellarocas2005, Judith2006}.
Time series-based detection constructs and monitors time series to spot such changes in review volume and rating.
A time series is a sequence of temporally ordered random variables $\bm{x}=[X_1, X_2, \dots, X_t, \dots]$,
and $\bm{x}_{m}^{n}=[X_m, \dots, X_n]$ denotes the portion
from time window $m$ to $n$.
For the $t$-th window (we also refer $t$ to the window or the timespan of the window),
the signals NR (number of reviews) and CAR (cumulative average rating) 
can be calculated to obtain two time series:
\begin{equation}
\mbox{NR}(t)=|\{r:t(r)\in t\}|,
\hspace{.1in}
\mbox{CAR}(t) = \frac{\sum_{t(r)\leq t} s(r)}{N_t},
\nonumber
\end{equation}
where $N_t$ is the number of reviews ever posted up to window $t$.
These two series can capture the large volume of spamming reviews
and inflated average ratings.
Changes in NR and CAR, denoted by $\Delta$NR and $\Delta$CAR,
can capture the abrupt changes in the volume of reviews and accumulated average rating:
\begin{equation}
\Delta\mbox{NR}(t)=\mbox{NR}(t) - \mbox{NR}(t-1),
\nonumber
\end{equation}
\begin{equation}
\Delta\mbox{CAR}(t) = \mbox{CAR}(t) - \mbox{CAR}(t-1).
\nonumber
\end{equation}

The deviation of the actual time series value
from the value predicted by a model that assumes smoothness of the series,
such as auto-regressive models,
can capture unexpected changes in the time series.
In particular, an order $d$ auto-regressive model (AR($d$)) predicts $X_t$
using historic data $\bm{x}_{t-d}^{t-1}$ and a linear model $\boldsymbol{\theta}^{(t)}$
\begin{equation}
X_t=\sum_{i=1}^{d}\theta_{i}^{(t)} X_{t-i}=\left< \boldsymbol{\theta}^{(t)}, \bm{x}_{t-d}^{t-1}\right>\label{eq:ar_d}.
\end{equation}
The deviation of the predicted CAR ($\widehat{\mbox{CAR}}(t)$)
from the actual CAR, can be used for detection (only promotion is considered):
\begin{equation}
\mbox{CAR-DEV}(t) = \max\{\widehat{\mbox{CAR}}(t) - \mbox{CAR}(t), 0\}.
\nonumber
\end{equation}
The larger the CAR-DEV, the more suspicious the window.

\subsection{Distribution-based detection signals}
\label{sec:dist_signals}
A spammer needs to post a large number of positive fake reviews
to promote the target.
Thus if the percentage of positive reviews within a window
is abnormally high, there are likely spamming activities.
The signal PR (Positive Ratio)~\cite{Rayana2015Collective, MukherjeeV0G13} is calculated based on this intuition:
\begin{equation}
\mbox{PR}(t) = \frac{|r:s(r)\geq 4\mbox{ and }t(r)\in t|}{n_t},
\nonumber
\end{equation}
where $n_t$ is the number of reviews within window $t$.
Second, the overall rating distributions of the $t$-th window
$\bm{p}(t)=[p_1(t),\dots, p_5(t)]$, with $p_i(t)$ be estimated by $|r:s(r)=i\mbox{ and } t(r)\in t|/n_t$,
can be perturbed by spamming ratings and deviate from the background rating distribution.
Such distortion in rating distribution
can be used as for spam detection~\cite{Feng2012, Minnich2015, Rayana2015Collective}.
Let $\bm{p}=[p_1,\dots, p_5]$ be the rating distribution of all historic ratings up to time $t$: $p_i=|r:s(r)=i\mbox{ and } t(r)\leq t|/N_t$.
The KL divergence between these two distributions
detects distortion in rating distribution:
\begin{equation}
\kldiv{\bm{p}(t)}{\bm{p}}=\sum_{i=1}^{5}p_i(t)\log \frac{p_i(t)}{p_i}
\nonumber
\end{equation}
The larger the KL-DIV, the farther $\bm{p}(t)$
is away from $\bm{p}$, and thus the more suspicious the $t$-th window.
Third, define the rating entropy 
\begin{equation}
\mbox{EN}(\bm{p}(t))=-\sum_{i=1}^{5}p_i(t)\log p_i(t)
\nonumber
\end{equation}
If the rating entropy of a window is low,
then the ratings therein are highly concentrating on a certain value
while a normal distribution shall have a certain level of dispersion across multiple values~\cite{Rayana2015Collective} (such as a U-shape~\cite{Hu2009}).
A related signal is the change in rating entropy
$\Delta\mbox{EN}=\mbox{EN}(t)-\mbox{EN}(t-1)$.
The window $t$ is suspicious if $\Delta$EN$<0$.

\section{EMERAL: an evasion generator}
\label{sec:attack}
After discussing the threat model,
we present EMERAL (Evasion via Maximum Entropy and Rating sAmpLing)
to generate evasions
against behavior-based detection signals.
The resulting optimization problem allows effective and efficient evasion generation (Section~\ref{sec:evasion_emp}).

\subsection{Threat model}
\begin{figure*}[t]
\centering     
\subfloat[Number of spams\label{fig:prop_a_amazon}]{\includegraphics[width=0.20\textwidth]{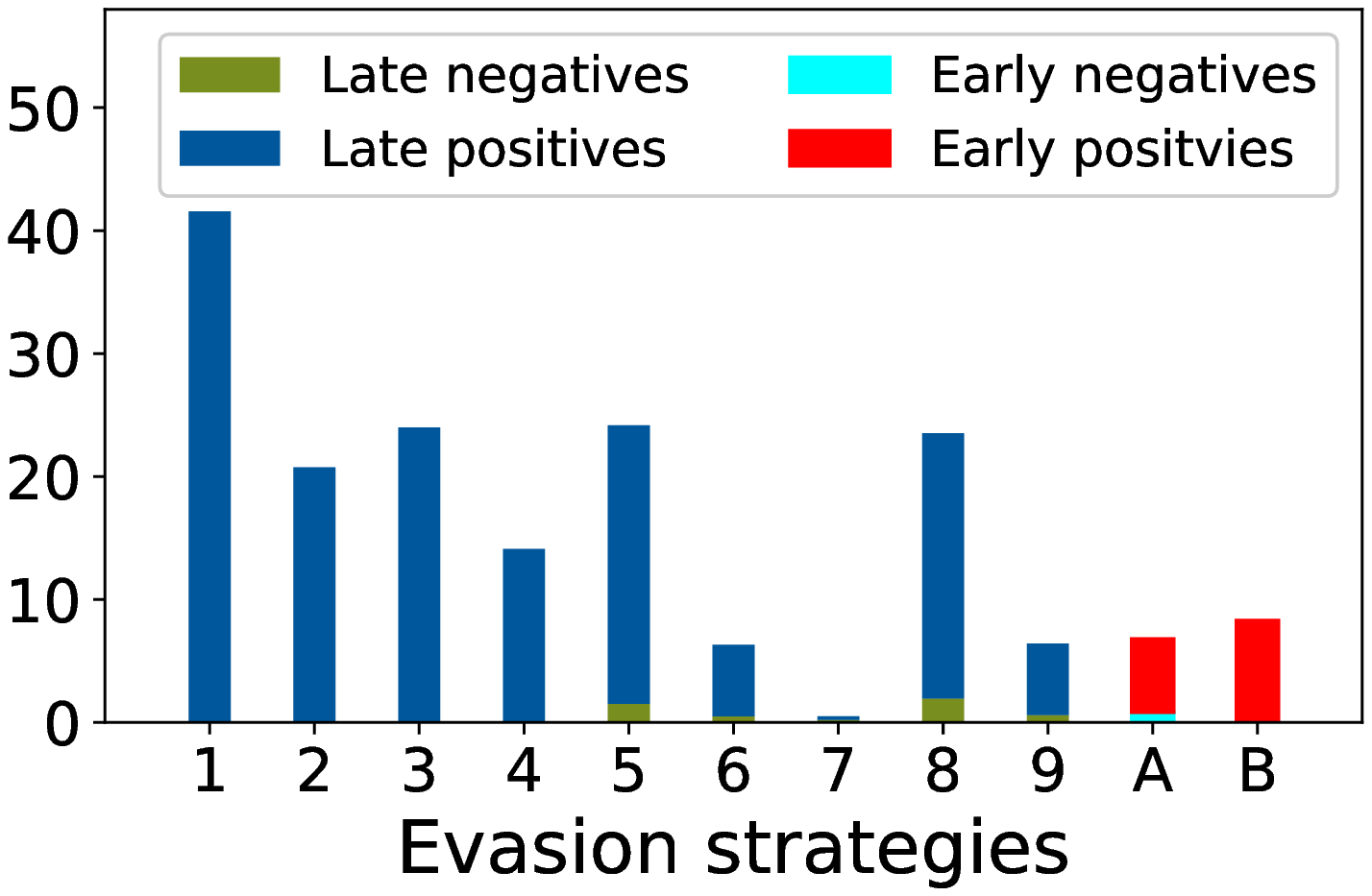}}%
\subfloat[Attack rates\label{fig:prop_b_amazon}]{\includegraphics[width=0.20\textwidth]{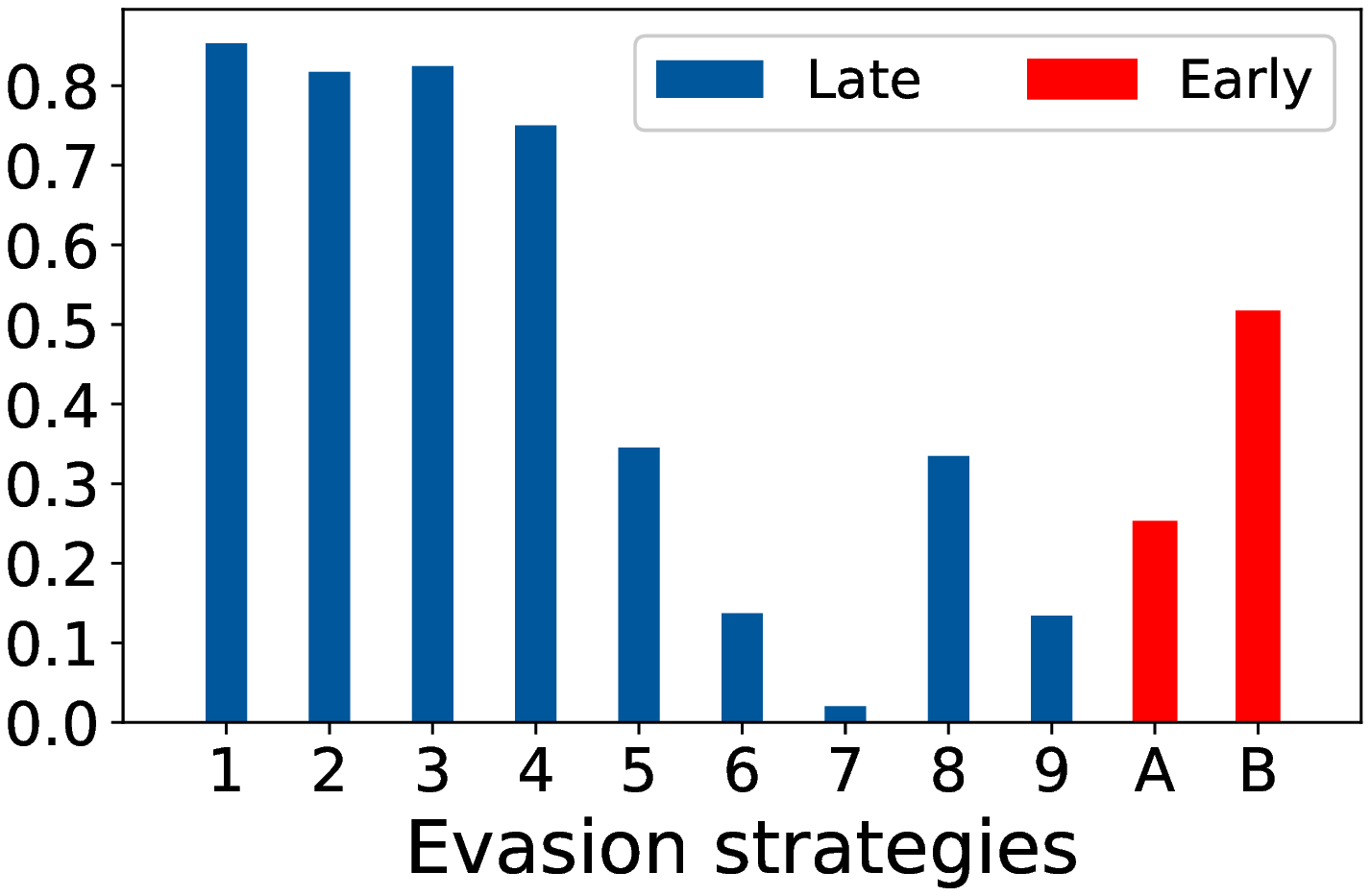}}%
\subfloat[Target CMR\label{fig:prop_c_amazon}]{\includegraphics[width=0.20\textwidth]{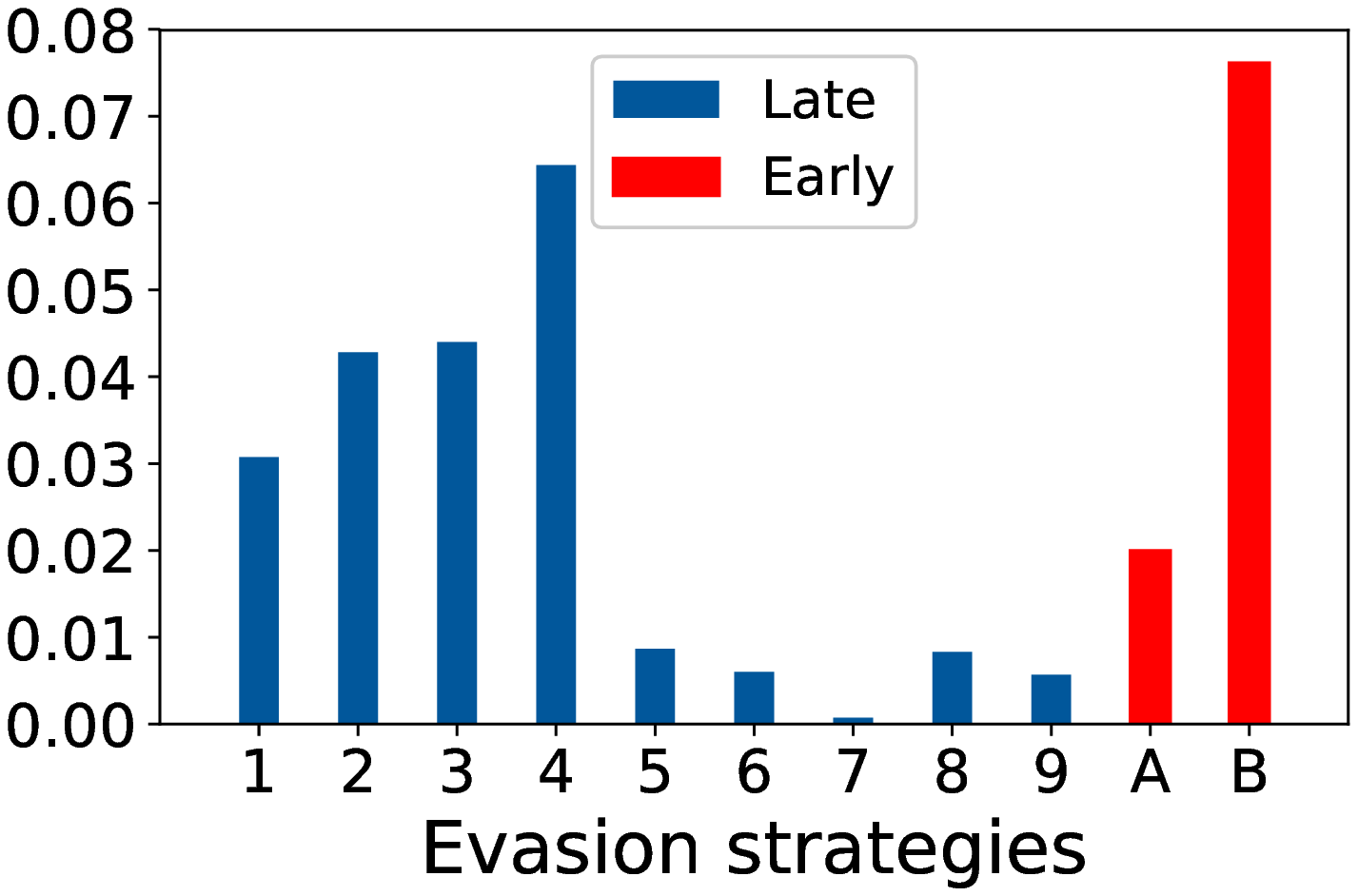}}%
\subfloat[Target CAR\label{fig:prop_d_amazon}]{\includegraphics[width=0.20\textwidth]{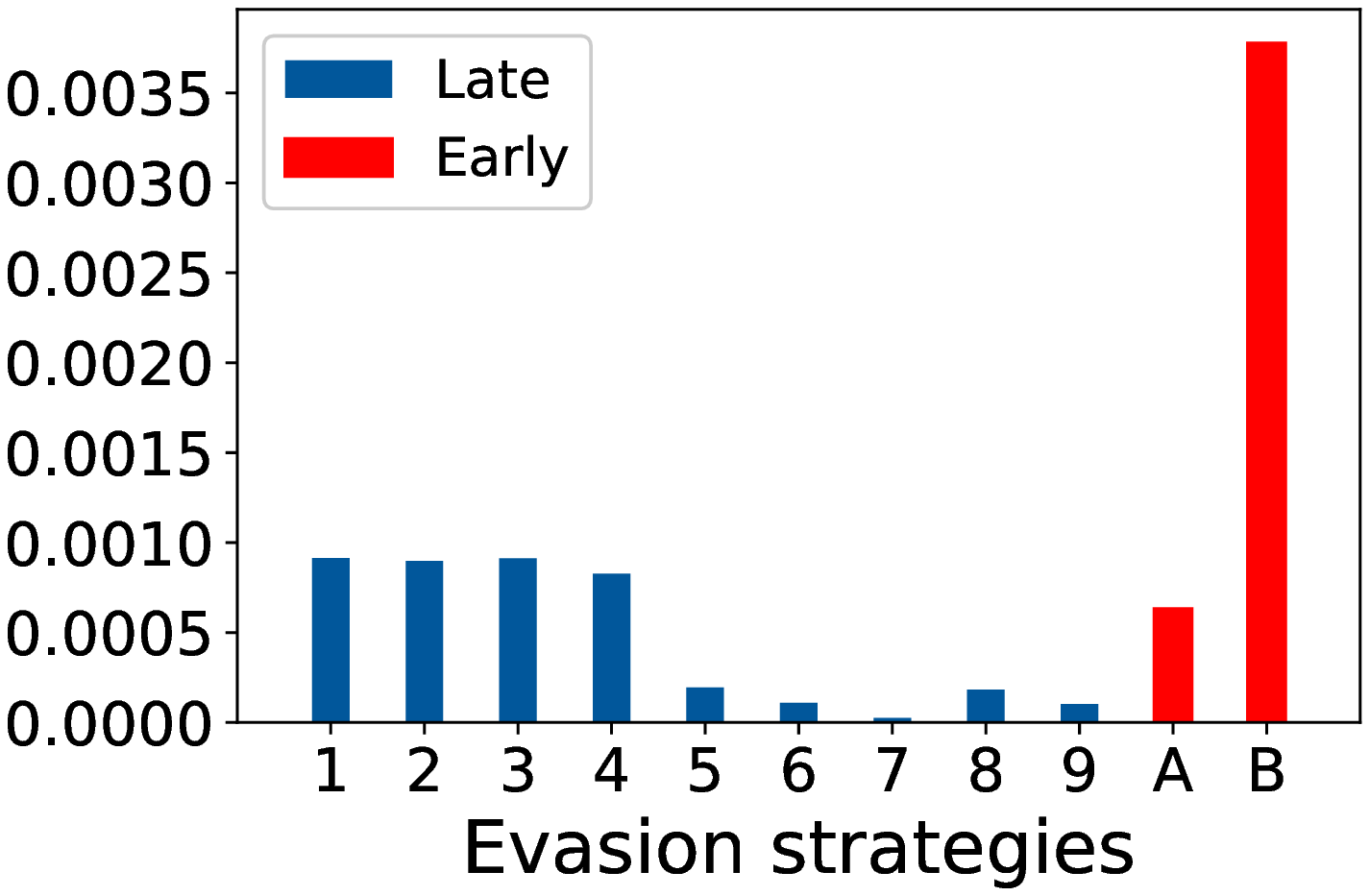}}%
\subfloat[Sensitivity to $p$\label{fig:sensitivity_p}]{\includegraphics[width=0.2\textwidth]{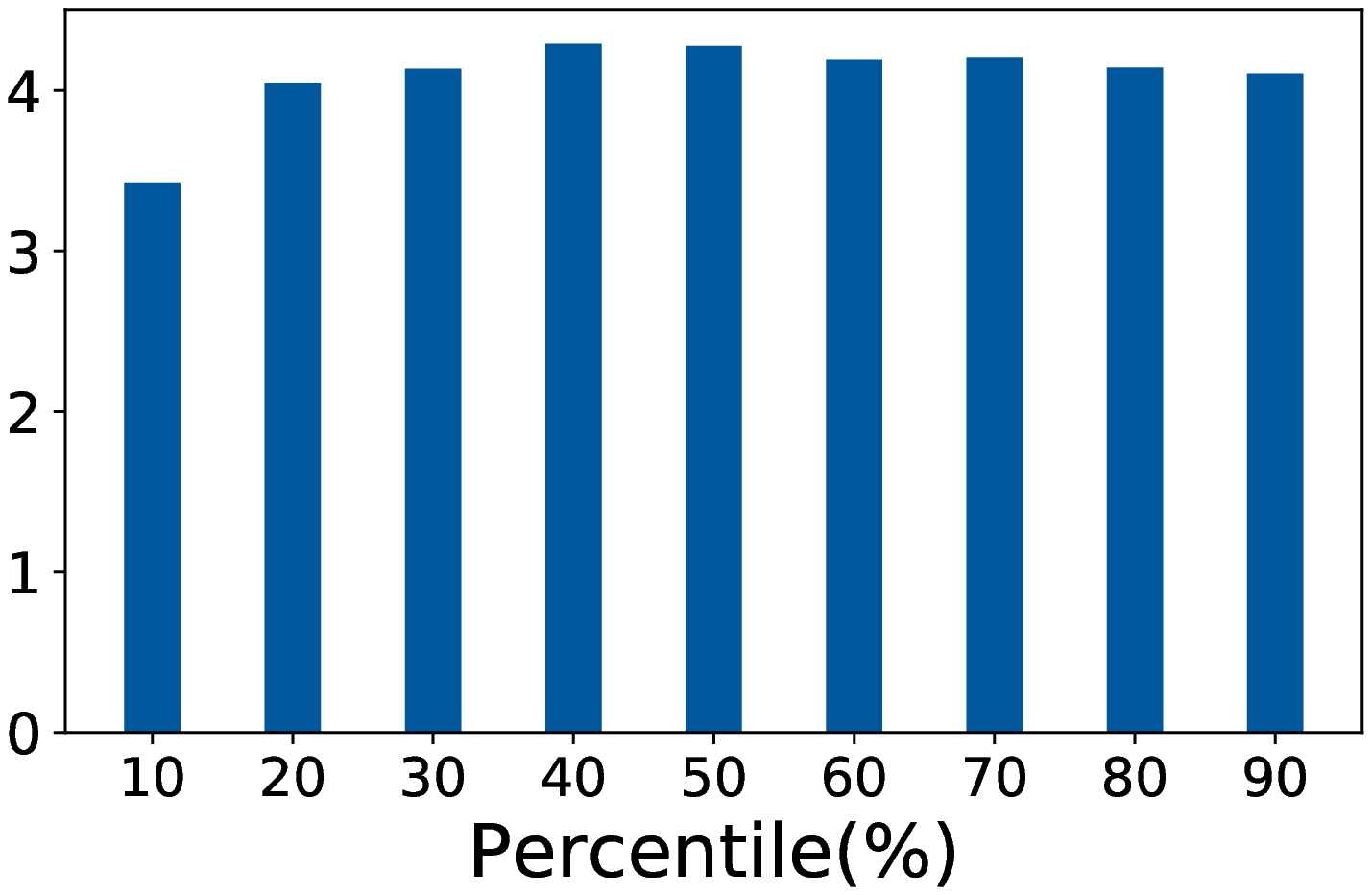}}
	\vspace{-0.5em}
\subfloat[Number of spams\label{fig:prop_a_yelp}]{\includegraphics[width=0.20\textwidth]{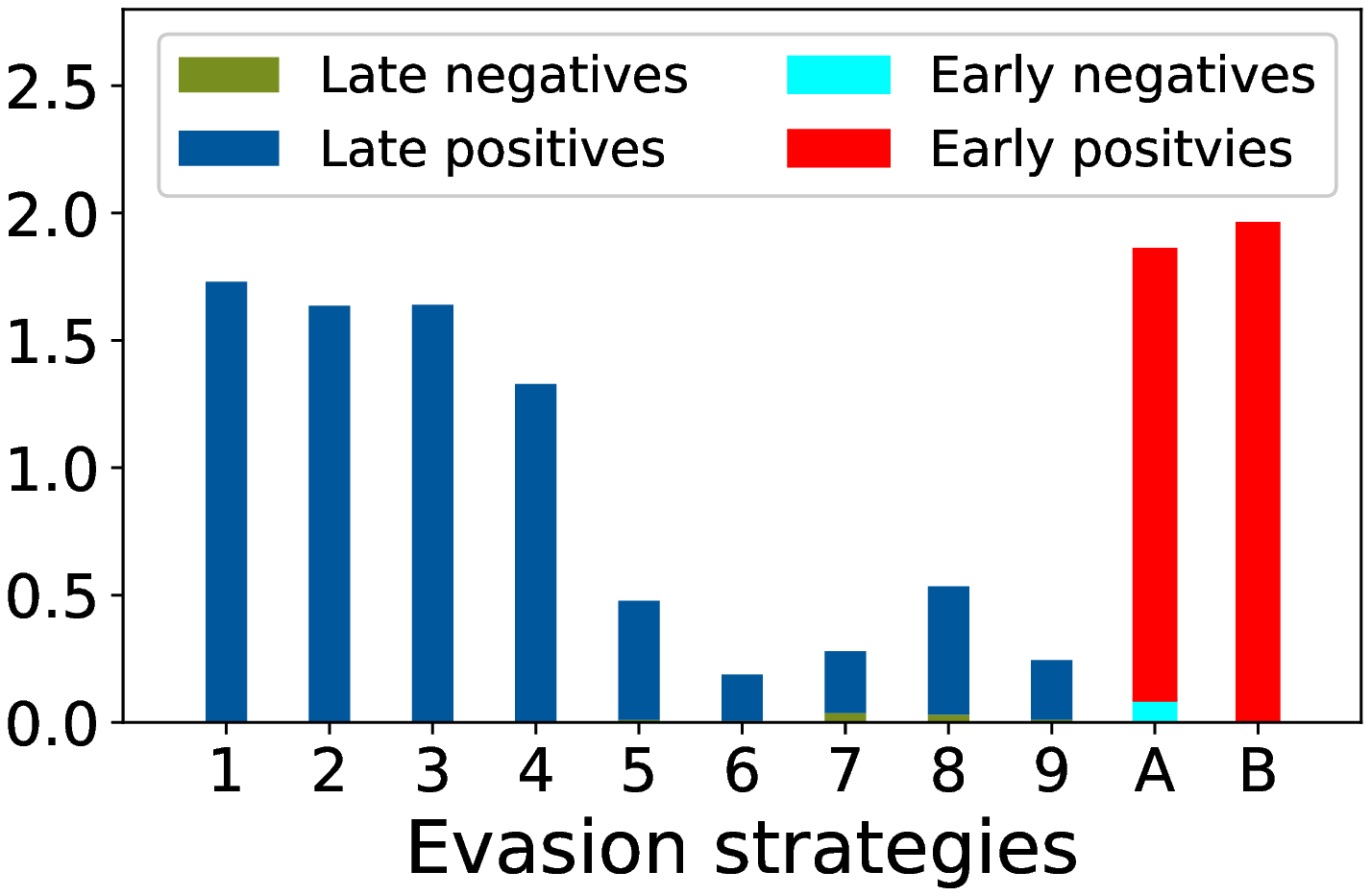}}%
\subfloat[Attack rates\label{fig:prop_b_yelp}]{\includegraphics[width=0.20\textwidth]{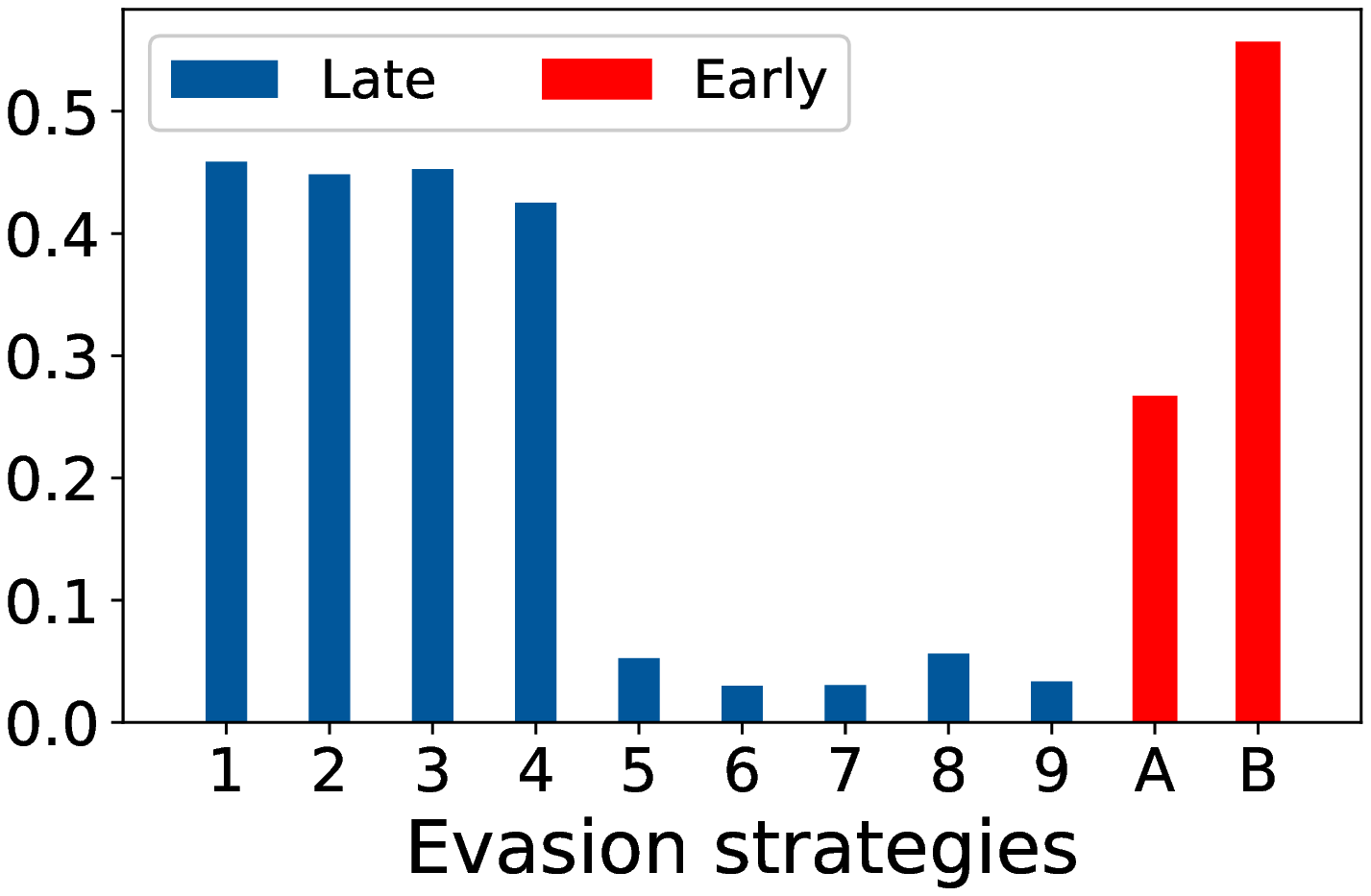}}%
\subfloat[Target CMR\label{fig:prop_c_yelp}]{\includegraphics[width=0.20\textwidth]{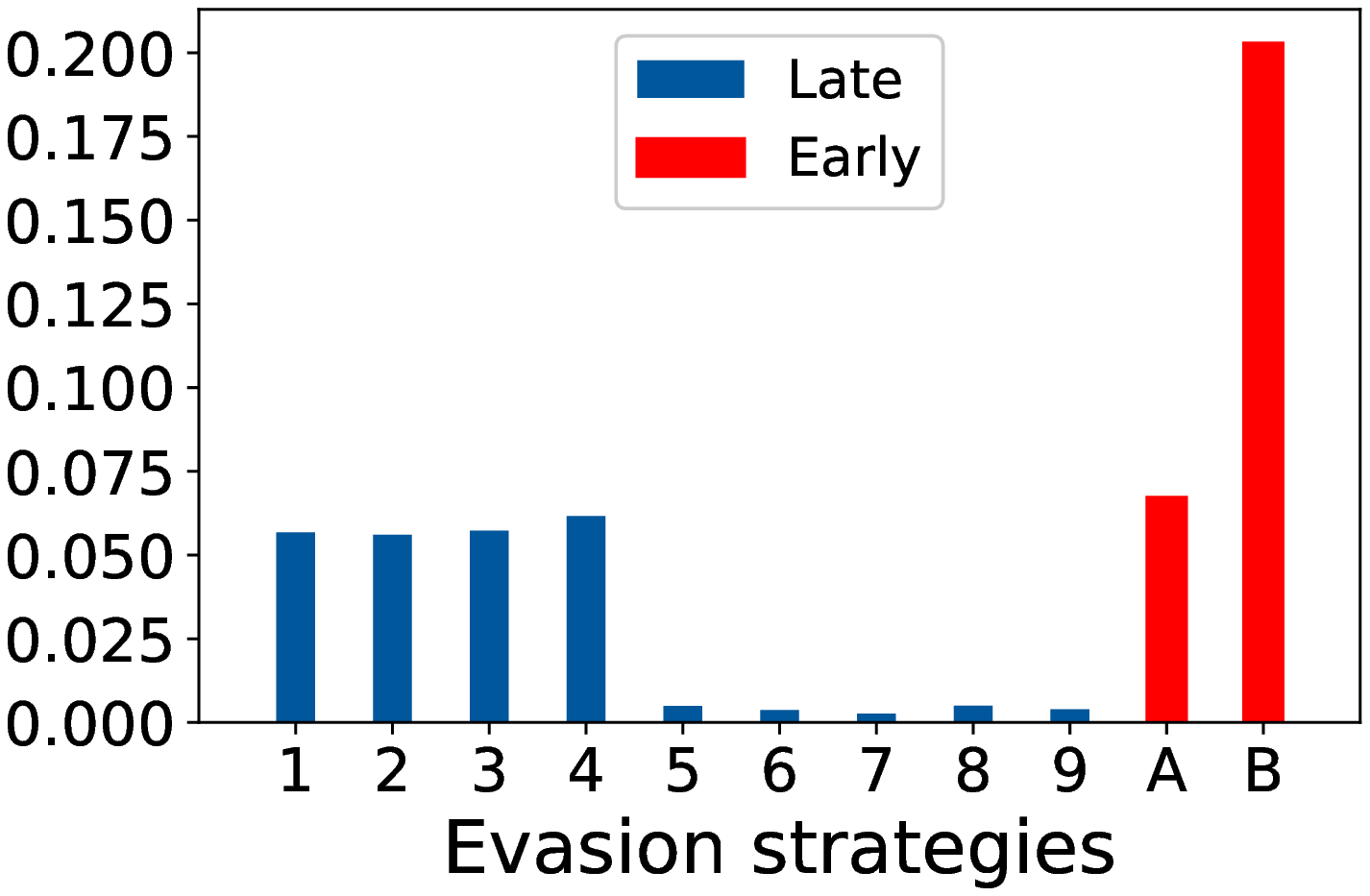}}%
\subfloat[Target CAR\label{fig:prop_d_yelp}]{\includegraphics[width=0.20\textwidth]{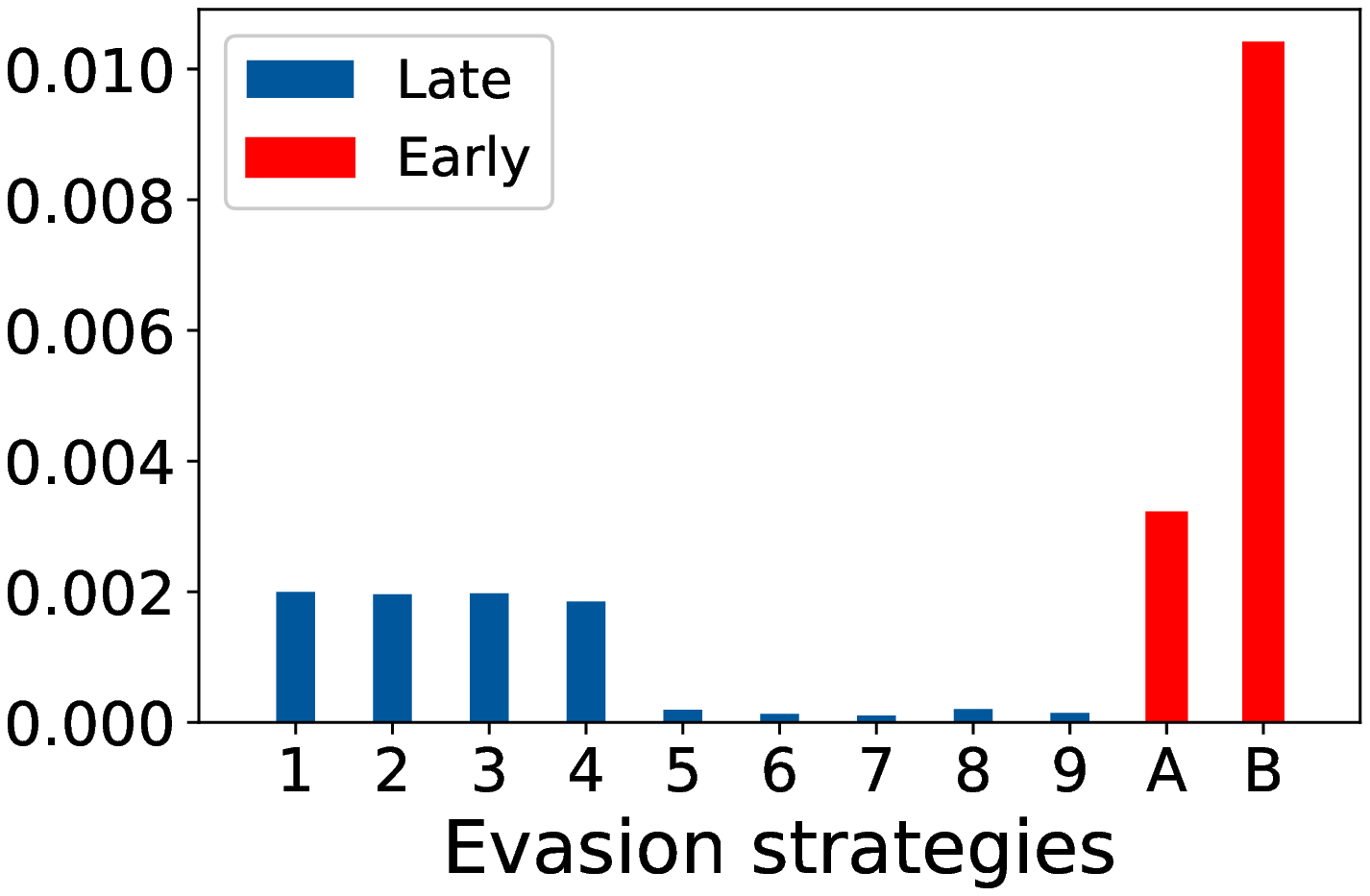}}%
\subfloat[Sensitivity to $d$\label{fig:sensitivity_d}]{\includegraphics[width=0.2\textwidth]{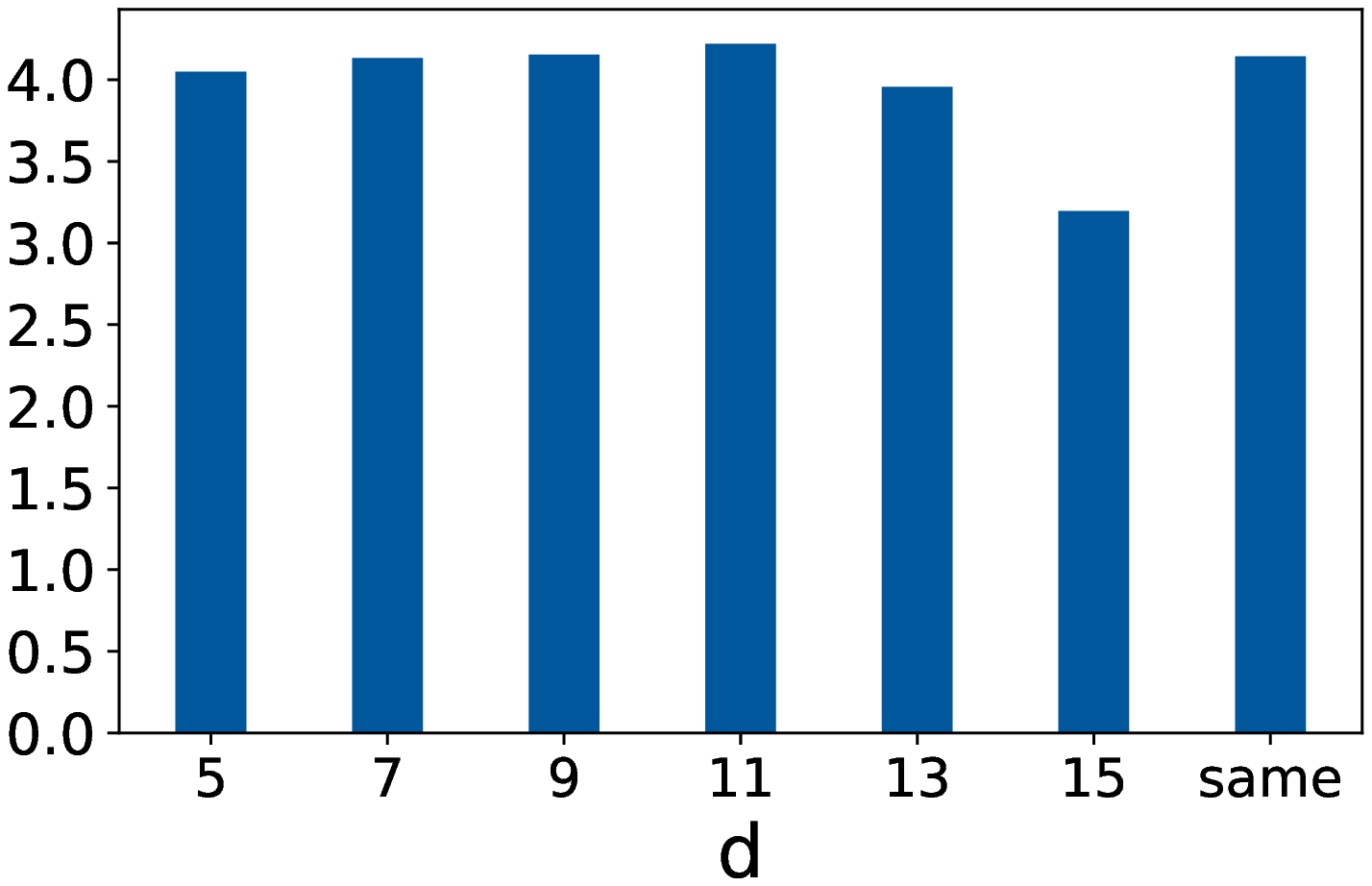}}
\caption{\small First 4 columns (top: Amazon, bottom: Yelp) show, from left to right:
the number of total/negative/positive spams posted,
successful rate of evasion,
average promotions in CMR and CAR under late and early evasions.
Last column: sensitivity of ranking promotion per spam on Amazon regarding
parameters $p$ and $d$ of the defender
(other metrics are similar on both datasets).
}
\label{fig:evasion_properties}
\vspace{-0.1in}
\end{figure*}

A threat model captures what knowledge about the defense system
an adversary can learn about and exploit to evade the defense system~\cite{Chen2017, liang2017, Vorobeychik2014, Srndic2014}.
Abundant review data,
including account and item profiles, review ratings and timestamps,
are publicly available on review websites to all users, including spammers.
For new or less popular products,
less historic data is available and yet they have a higher incentive to spam.
An evasion should be able to generate attacks even with scarce data.
Obtaining labeled data is easy through multiple channels:
1) released review data are filtered before being made public and thus represent normal reviews;
2) Yelp further releases identified spams;
3) spam spotting services, such as Fakespot and ReviewMeta,
release predicted class labels or probability too.
Behavior-based detection signals are published with great details~\cite{MukherjeeV0G13}.
Fakespot and ReviewMeta further explain to users what detection signals are used to detec spams.
Regarding hyper-parameter for detection signal constructions,
we empirically show that the proposed evasion model does not require exact knowledge for effective evasions (Figures~\ref{fig:sensitivity_p} and~\ref{fig:sensitivity_d}).
Signals based on review texts or graphs are orthogonal to the behavior-based signals,
and a spammer does not need knowledge about these signals
to conduct successful evasions.
A detection algorithm essentially aggregate multiple detection signals for detection.
Spammers can have different levels of knowledge about the aggregation.
A naive spammer can only be aware of the signals but not how they are aggregated.
A spammer with moderate knowledge can know and assume simple aggregation methods,
such as uniform linear combination or taking the most suspicious signal.
Lastly, a spammer with perfect knowledge knows the algorithm that aggregate the signals.

\subsection{Evading behavior-based signals}
\label{sec:emeral}
In a spamming campaign,
a spammer needs to know the exact ratings of each of spams
to manipulate target rating while evading signals based on rating distribution, such as KL-DIV, EN, $\Delta$EN and PR.
We propose to first find an evading rating distribution and then sample ratings from the distribution for the spams.

To evade KL-DIV,
all ratings, including spamming and normal ones,
in the current time window should
have a rating distribution $\bm{p}$ close
to $\bar{\bm{p}}$ that the defender considers normal.
Specifically, let $R\in\{1,2,\dots, 5\}$ be a random variable of ratings
such that $p(R=r)=p_r\geq 0$ and $\sum_{r=1}^{5}p_r=1$.
For a target with many ratings,
the spammer can estimate $\bar{p}_i$ from the ratings using MLE.
For the $t$-th window,
the spammer can find $\bm{p}$ with minimal KL-divergence to $\bar{\bm{p}}$
\begin{equation}
\begin{array}{ll}
\underset{\bm{p}}{\textnormal{min}}
& \textnormal{KL}(\bm{p} || \bm{\bar p}) = \sum_{r=1}^{5} p_r\log\frac{p_r}{\bar{p}_r}.
\end{array}
\end{equation}
The spammer wants to move CAR to $\tilde{x}_t$ from $x_t$, where
$x_{t}$ is the CAR at time $t$ without spams.
When targeting at promotional spamming,
the spammer needs the manipulated CAR
to be close to but not to exceed the target $\tilde{x}_t$.
Let
$N_{t-1}$ be the number of ratings accumulated up to time $t$,
$n_t$ be the number of existing ratings at time $t$
without the spamming ratings,
and $n_\delta$ be the number of spamming ratings to be added.
$\mathbb{E}_{\bm{p}}(R) = \sum_{r=1}^{5} r\times p_r$ is the expectation of $R$.
Then the manipulated CAR after the attack is
$(N_{t-1} x_{t-1} + (n_t + n_\delta) \mathbb{E}_{\bm{p}}[R]) / (N_t+n_t+n_{\delta})$
and the goal becomes:
\begin{equation}
\label{eq:car_range}
\tilde{x}_t - \epsilon
\leq \frac{N_{t-1} x_{t-1} + (n_t + n_\delta) \mathbb{E}_{\bm{p}}[R]}{N_t+n_t+n_{\delta}}
\leq \tilde{x}_t,
\end{equation}
where $\epsilon > 0$ is a small positive number to allow slack in $\tilde{x}_t$.
In addition,
the spammer can evade $\Delta$EN and NPR
by adding relevant constraints, leading to the
following inequality-constrained KL-divergence minimization problem:
\begin{equation}
\small
\label{eq:kl_primal}
\begin{array}{ll}
\underset{\bm{p}}{\textnormal{min}}
& \textnormal{KL}(\bm{p}||\bar{\bm{p}})\\[1em]
\textnormal{s.t.}
&\mathbb{E}_{\bm{p}}[R] \leq U \triangleq \dfrac{(N_t + n_t + n_\delta) \tilde{x}_t - N_{t-1} x_{t-1}}{n_t + n_\delta}, \\[1em]
&-\mathbb{E}_{\bm{p}}[R] \leq B \triangleq \dfrac{(N_t + n_t + n_\delta) (\tilde{x}_t - \epsilon) - N_{t-1} x_{t-1}}{n_t + n_\delta}, \\[1em]
& -{H}(\bm{p}) \leq -(H_{t-1}+H_\delta) \triangleq - H, \\[1em]
& p_4+p_5 \leq P, \hspace{.3in} \sum_r p_r = 1.
\end{array}
\end{equation}
The first two constraints are derived from Eq.~(\ref{eq:car_range}),
and the third enforces the rating distribution entropy $H(\bm{p})$ to be at least
$H_{t-1} + H_{\delta}$ to evade $\Delta$EN (change in entropy).
The constraint $p_4 + p_5 < P$ ensures that after spamming,
the ratio of positive reviews (4 and 5 star ratings)
will not exceed $P$ to evade PR (ratio of positives).
The optimization
can be solved using Lagrangian multiplier method:
\begin{tcolorbox}[width=3.45in,
boxsep=0pt,
left=1pt,
right=16pt,
top=2pt,
bottom=0pt,
arc=0pt,
boxrule=1pt,
toprule=1pt,
colback=white
]
\begin{equation}
\label{eq:kl_dual}
\begin{aligned}
& \underset{\alpha,\beta,\gamma,\lambda}{\max} 
& & L(\alpha,\beta,\gamma,\lambda) \\
& \text{s.t.}
& & \alpha \geq 0, \beta \geq 0, \gamma \geq 0, \lambda \geq 0
\end{aligned}
\end{equation}
\end{tcolorbox}
\vspace{-0.2in}
\begin{equation}
L(\alpha, \beta, \lambda, \gamma)=-(1+\gamma)\log{Z}-(1+\gamma)-\alpha U+\beta B-\lambda p+\gamma H \nonumber
\end{equation}
and
$Z=\sum_{r=1}^{5} p_r=\sum_{r=1}^{5}\exp(S_r / (1+\gamma))$ with 
\begin{equation}
S_r=\log{\bar{p}_r}-(\alpha-\beta)r-1 - \lambda \mathbb{I}(r)-\gamma
\end{equation}
We can use gradient ascent to find the optimal Lagrangian multipliers $\alpha^{\ast}$, $\beta^{\ast}$, $\gamma^{\ast}$ and $\lambda^{\ast}$.
Evading EN is similar and can be done by
setting the target distribution $\bar{\bm{p}}$ to the uniform distribution.

The above optimization problem assumes that the number of spamming reviews ($n_\delta)$
and the target CAR value ($\tilde{x}_t$) are given.
We further set these parameters to evade $\Delta$CAR (change in Cumulative Average Rating),
CAR-DEV (Cumulative Average Rating deviation)
and $\Delta$NR (change in the number of reviews)
that focus on abrupt changes in time series.
By assuming that the defender adopts a degree $d$ AR model $\boldsymbol{\theta}$ to capture CAR deviation,
the spammer sets $\delta_t=\tilde{x}_t - x_t$ so that
1) $\tilde{x}_t$ is as high as possible;
2) $|\tilde{x}_t - \hat{x}_t|<\epsilon$ to evade the detection of CAR-DEV;
where $\epsilon$ is a small number and $\hat{x}_t$ is the predicted CAR by the AR model.
3) $\hat{x}_{t+1}(\delta_t)$ is maximized to
allow a larger $\delta_{t+1}$ to be added to $x_{t+1}$
while $|x_{t+1}(\delta_t) + \delta_{t+1} - \hat{x}_{t+1}|<\epsilon$.

Note that $\hat{x}_{t+1}$ is a function of $\delta_t$ since
the next AR model $\boldsymbol{\theta}^{(t+1)}$ is updated on $\tilde{x}_t$.
Overall the spammer can
 \vspace{-0.2em}
\begin{eqnarray}
\label{eq:evade_car_dev}
\begin{aligned}
& \underset{\delta}{\textnormal{max}}
& & \delta +  \hat{x}_{t+1}(\delta)\label{eq:evade_delta_car_1}\\
& \textnormal{s.t.}
& & 0\leq \delta,
\hspace{.1in}
|\hat{x}_t - (x_t + \delta)| < \epsilon,
\hspace{.1in}
x_t + \delta < U,\\
\end{aligned}
\end{eqnarray}
where $U$ is an upper bound of the time series ($U=5$ for CAR).
Assuming the spammer mimics the defender by 
training $\boldsymbol{\theta}$ using online gradient descent with learning rate $\eta$,
then
Eq.~(\ref{eq:evade_delta_car_1}) becomes the following constrained quadratic programming problem:
\begin{tcolorbox}[width=3.5in,
boxsep=0pt,
left=1pt,
right=0pt,
top=2pt,
arc=0pt,
boxrule=1pt,
toprule=1pt,
colback=white
]
\begin{equation}
\begin{array}{ll}
\underset{\delta}{\textnormal{max}}
&\left[1+\theta^{(t+1)}_{1} + \eta \left({\bm{x}_{t-d}^{t-1}}\right)^{\top} \bm{x}_{t-d+1}^{t}\right]\delta + \eta x_{t-1} \delta^2 \label{eq:opt_delta}\\
\textnormal{s.t.}
& \max\{0,\hat{x}_t-x_t-\epsilon\} \leq \delta \\
& \min\{U-x_t, \hat{x}_t-x_t+\epsilon\} \geq \delta \nonumber
\end{array}
\end{equation}
\end{tcolorbox}
The optimal $\delta$ is denoted by $\delta_{t}^{\ast}$
and is used to set $\tilde{x}_t=x_t+\delta_t^{\ast}$ in Eq.~(\ref{eq:car_range}).
The spammer also wants to evade detection based on burst detection~\cite{Xie2012a, Fei13}.
We can add the constraint $|x_{t-1} - (x_t + \delta)| < \epsilon$,
to reduce $\Delta$CAR.
To reduce $\Delta$NR,
a spammer samples 
$n_\delta$ ratings from the distribution obtained from Eq.~(\ref{eq:kl_primal}),
such that $n_\delta$ is below the $p$-percentile of
all positive historical increments in NR.
To optimize $\delta$, the spammer needs to know both the degree of AR model
and the CDF (cumulative distribution function) of the detection signals on the defender side.
We empirically show that lacking knowledge of $d$ and $p$
will not prevent spammers from conducting effective and evasive attacks.

\begin{algorithm}
\small
\caption{EMERAL}
\begin{algorithmic}
\STATE {\bf Input}: Reviews of a target; maximum number of trials $M$.
\STATE {\bf Output}: Ratings of spamming reviews to be posted.
\STATE Select $n_\delta$ and $\delta_t^{\ast}$ based on historic reviews.
\STATE Set rating distribution: $p_5=1$ and $p_i=0,i=1,\dots,4$.
\IF{Evade rating distribution-based signals}
\STATE Use $n_\delta$ and $\delta_t^{\ast}$
to solve problem (\ref{eq:kl_primal}) to find $\bm{p}(t)$.
\STATE Exit without spams if no evasive rating distribution found.
\ENDIF
\WHILE{Not succeed and number of trials $<M$}
\STATE Sample $n_\delta$ ratings from $\bm{p}(t)$ satisfying constraints $\delta_t^\ast$.
\STATE Return sampled ratings if no constraint violated.
\ENDWHILE
\STATE Exit without spams.
\end{algorithmic}
\label{alg:evasion_alg}
\end{algorithm}

The overall evasion procedure EMERAL is described in Algorithm~\ref{alg:evasion_alg}.
EMERAL requires the target to have a reasonably long history of reviews to calculate the evasion parameters.
Note that the algorithm may fail to find an evasive spamming plan for a window, and in that case, the spammer will not attempt to attack.
Based on preliminary experiment, there are only 
9 combinations of detection singals,
denoted by E1 to E9~\footnote{
	The 9 combinations are:
E1=[NR],
E2=[NR, $\Delta$NR],
E3=[NR, CAR-DEV],
E4=[NR, CAR-DEV, $\Delta$CAR],
E5=[NR, CAR-DEV, KL-DIV],
E6=[NR, CAR-DEV, KL-DIV, $\Delta$EN],
E7=[NR, CAR-DEV, KL-DIV, NPR],
E8=[NR, CAR-DEV, EN],
E9=[NR, CAR-DEV, EN, $\Delta$EN]
}, that are profitable for the spammers to evade.


\subsection{EMERAL for early spamming: evasion E-A and E-B}
\label{sec:early_evasion}
It is shown that dishonest businesses have a strong motivation to conduct promotional spamming
early on when their products are open for review~\cite{Santosh2016, Rayana2015Collective}.
We adapt EMERAL to generate evasive spams for such situations.
A new product will have a smaller number of reviews for a spammer
to probe evasion parameters from the CDFs of the signals.
However, a spammer can leverage the CDFs of the signals based on the early reviews of other products and estimate the evasion parameters.
In particular, a spammer can obtain NR, $\Delta$NR, $\Delta$CAR
and rating distribution
of the early time windows of all available products,
and then tries 
maximize the entropy
while safisfying constraints over NR, $\Delta$NR and $\Delta$CAR (E-A).
Evasion E-B tries to post a maximum number of 5 star reviews to evade
NR, $\Delta$NR and $\Delta$CAR at the same time.

\subsection{Empirical properties of EMERAL on late spamming}
\label{sec:evasion_emp}

We use datasets collected from Amazon and Yelp~\cite{He2016, Rayana2015Collective}.
To spam targets with long review histories,
we filter products on Amazon with less than 1000 reviews or having less than 37 weeks of reviews,
and restaurants on Yelp having less than 37 months of reviews (a month/week is referred to as a ``time window'' on the two datasets, respectively).
The results are 383 products with 1175088 reviews on Amazon, and 327 restaurants with 247117 reviews on Yelp.
The evasions are created on each target
for the last 5 consecutive time windows
based on knowledge obtained from all previous time windows (32 in total).
We compute evasions with strategies E1 to E9,
assuming that the spammer aims to keep each detection signal
lower than the 80 percentiles of the corresponding signals' CDFs after the attacks.

The average numbers of total/negative/positive spams
posted in all test windows by each evasion on the two datasets
are shown in Figure~\ref{fig:prop_a_amazon} and~\ref{fig:prop_a_yelp}.
One can observe that all evasions post
much more positive spams than negatives
to promote business ratings and rankings.
Interestingly,
if a spammer decides to evade rating distribution related signals,
as with evasions E5 to E9,
some negative reviews have to be posted,
while with evasions E1 to E4, there is no negative reviews.
Since EMERAL
does not guarantee that an evasive rating distribution can be found,
Figures~\ref{fig:prop_b_amazon} and~\ref{fig:prop_b_yelp}
show the percentages of windows that an evasion is possible.
Evasions 1-4 are successful in most of the windows
(more than 70\%) while Evasions 5-9 are more conservative due to constraints over rating distribution.
Figures~\ref{fig:prop_c_amazon} and~\ref{fig:prop_d_amazon} on the top row
show the promotions in the target's CMR and CAR per spamming review,
averaged over all targets and test windows, on the Amazon dataset.
We can see that evasions 5-9 are less profitable to the spammers as the promotions are rather small,
and evasions 1-4 can promote the target rather effectively.
We tried to evade other combinations of the signals using EMERAL but found out that it is hard to find evasions valuable to the spammers.
As a result, the defender needs not to consider evasions against other combination of signals in Table~\ref{tab:detection_signals}.

\subsection{Empirical properties of EMERAL on early spamming}
The early windows of the datasets that are not used for late spamming are used for early evasions.
In the same set of figures for late spamming,
we use the last two bars in each subfigure
to demonstrate the properties of early evasive spamming.
In Figures~\ref{fig:prop_a_amazon} and~\ref{fig:prop_a_yelp},
we can see that average numbers of total/negative/positive spams post in each early windows.
There is not large difference in the total number,
but E-A creates a small amount of negative reviews
due to entropy maximization.
In Figures~\ref{fig:prop_b_amazon} and~\ref{fig:prop_b_yelp},
we can see that E-B has a successful rate two times of the rate of E-A,
leading to higher per spam utility in CMR and CAR promotions,
shown in Figures~\ref{fig:prop_c_amazon}, \ref{fig:prop_c_yelp},
\ref{fig:prop_d_amazon} and \ref{fig:prop_d_yelp}.
We conclude that early spamming is very attractive to spammers
and advanced defense against early spamming need to be deployed, as we will do next.

\textbf{Spammer knowledge requirements}
From Section~\ref{sec:emeral},
for evasive late spamming, it seems that the spammer needs to know
the degree the AR model and the $p$-percentile of the CDF of historic CAR to find out $\epsilon$ for solving Eq.~(\ref{eq:opt_delta}).
For evasion E4, which requires $d$ and $p$,
the two parameters can be selected from wide ranges so that they can be different from the values used by the defender.
Figures~\ref{fig:sensitivity_p} and~\ref{fig:sensitivity_d}
show the ranking promotion brought by E4 after detection based on the spammers' assumptions 
is not much affected by the inaccurate knowledge of these two hyper-parameters.
For example,
different $d$ values achieve similar spamming utility as when $d$ is the same as the value set by the defender.
There is no parameter $d$ in early spamming.

\section{DETER: Evasion agnostic defenses}
\label{sec:new_defense}

\begin{table*}[t]
\footnotesize
\setlength\tabcolsep{3pt}
\caption{
Detection AUC under different strategy profiles with late/early spamming. Top: Amazon, buttom: Yelp.}
\label{tab:all_defense}

\subfloat[AUC of detection of late spamming on Amazon (rows: evasions, columns: defenses). \label{tab:late_auc_amazon}]{
\begin{tabular}{l || cccc|ccc}
\toprule
 Evasions & Best\_P & $\mathbf{w}^m$ & $\mathbf{w}^{a}$ & $\mathbf{w}^r$ & EN\_M & EN\_A & DETER\\\hline
Rand1& 0.91& 0.91$_{\pm 0.007}$ & 0.91$_{\pm 0.007}$ & 0.89$_{\pm 0.008}$ & 0.74$_{\pm 0.007}$ & 0.90& \textbf{0.93}$_{\pm 0.006}$\\
Rand2& 0.86& 0.84& 0.84& 0.83$_{\pm 0.006}$ & 0.73$_{\pm 0.011}$ & 0.84$_{\pm 0.006}$ & \textbf{0.87}$_{\pm 0.006}$\\\hline
E1& 0.91& 0.85& 0.86& 0.84$_{\pm 0.007}$ & 0.91& 0.89& \textbf{0.92}\\
E2& 0.90& 0.84& 0.85& 0.83& \textbf{0.91}& 0.88& 0.90\\
E3& \textbf{0.89}$^{\ast}$& 0.82& 0.83& 0.81$_{\pm 0.008}$ & 0.87& 0.85& 0.88\\
E4& \textbf{0.75}$^{\ast}$& 0.67& 0.69& 0.67& 0.71& 0.72& 0.74\\\hline
E5& 0.97& 0.96& 0.95& 0.94& 0.71& 0.97& \textbf{0.98}\\
E6& 0.97& 0.93& 0.92& 0.90$_{\pm 0.007}$ & 0.62& 0.97& \textbf{0.98}\\
E7& \textbf{0.98}$^{\ast}$& 0.96$_{\pm 0.005}$ & 0.96& 0.95$_{\pm 0.011}$ & 0.72$_{\pm 0.011}$ & 0.97$_{\pm 0.007}$ & 0.98\\
E8& 0.97& 0.96& 0.96& 0.94& 0.72& 0.97& \textbf{0.98}\\
E9& \textbf{0.96}$^{\ast}$& 0.92& 0.91& 0.89$_{\pm 0.014}$ & 0.52$_{\pm 0.009}$ & 0.92$_{\pm 0.006}$ & 0.96\\
\bottomrule
\end{tabular}
}
\hfill
\subfloat[AUC of detection of early spamming on Amazon\label{fig:early_auc_amazon}]{
	\includegraphics[width=0.32\textwidth]{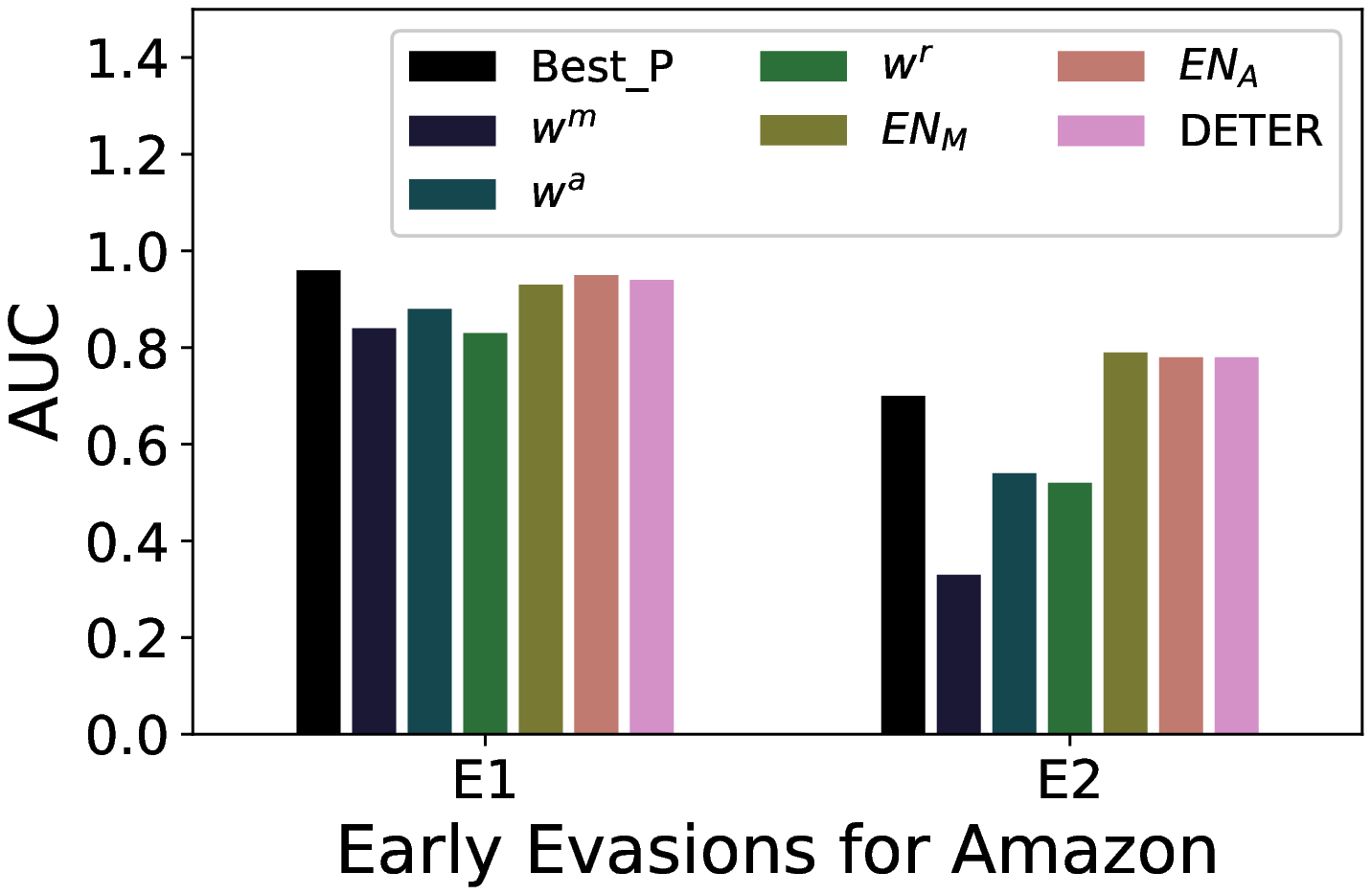}
	
}
\bigskip

\vspace{-.1in}

\subfloat[AUC of detection of late spamming on Yelp (rows: evasions, columns: defenses). \label{tab:late_auc_yelp}]{
\begin{tabular}{l || cccc|ccc}
\toprule
 Evasions & Best\_P & $\mathbf{w}^m$ & $\mathbf{w}^{a}$ & $\mathbf{w}^r$ & EN\_M & EN\_A & DETER \\\hline
Rand1& 0.77$_{\pm 0.012}$ & 0.71$_{\pm 0.012}$ & 0.73$_{\pm 0.011}$ & 0.71$_{\pm 0.016}$ & 0.67$_{\pm 0.015}$ & 0.70$_{\pm 0.017}$ & \textbf{0.80}$_{\pm 0.015}$ \\
Rand2& 0.73$_{\pm 0.013}$ & 0.65$_{\pm 0.012}$ & 0.67$_{\pm 0.012}$ & 0.65$_{\pm 0.016}$ & 0.73$_{\pm 0.044}$ & 0.73$_{\pm 0.051}$ & \textbf{0.76}$_{\pm 0.016}$ \\ \hline
E1 & 0.73& 0.63& 0.66& 0.65$_{\pm 0.007}$ & \textbf{0.82}& 0.81& 0.78\\                                                              
E2 & 0.73& 0.63& 0.66& 0.65& \textbf{0.82}& 0.81& 0.78\\                                                                             
E3 & 0.73& 0.63& 0.65& 0.64$_{\pm 0.007}$ & \textbf{0.81}& 0.80& 0.77\\                                                              
E4 & 0.69& 0.58& 0.61& 0.59$_{\pm 0.006}$ & \textbf{0.78}& 0.77& 0.71\\                                                               \hline
E5 & \textbf{0.99}$^{\ast}$& 0.97& 0.96& 0.95$_{\pm 0.011}$ & 0.71$_{\pm 0.009}$ & 0.93$_{\pm 0.009}$ & {}0.95$_{\pm 0.007}$ \\        
E6 & \textbf{0.99}$^{\ast}$& 0.95$_{\pm 0.008}$ & 0.95$_{\pm 0.008}$ & 0.94$_{\pm 0.008}$ & 0.75$_{\pm 0.007}$ & 0.93$_{\pm 0.007}$ & 0.97\\
E7 & \textbf{0.99}$^{\ast}$& 0.95& 0.95& 0.93$_{\pm 0.009}$ & 0.73& 0.95& 0.98\\                                                     
E8 & \textbf{0.99}$^{\ast}$& 0.96& 0.96& 0.95$_{\pm 0.012}$ & 0.74$_{\pm 0.008}$ & 0.95$_{\pm 0.007}$ & 0.97\\                       
E9 & \textbf{0.99}$^{\ast}$& 0.94& 0.94& 0.92$_{\pm 0.019}$ & 0.70$_{\pm 0.012}$ & 0.90$_{\pm 0.007}$ & 0.95\\                       
\bottomrule
\end{tabular}
}
\hfill
\subfloat[AUC of detection of early spamming on Yelp\label{fig:early_auc_yelp}]{
	\includegraphics[width=0.32\textwidth]{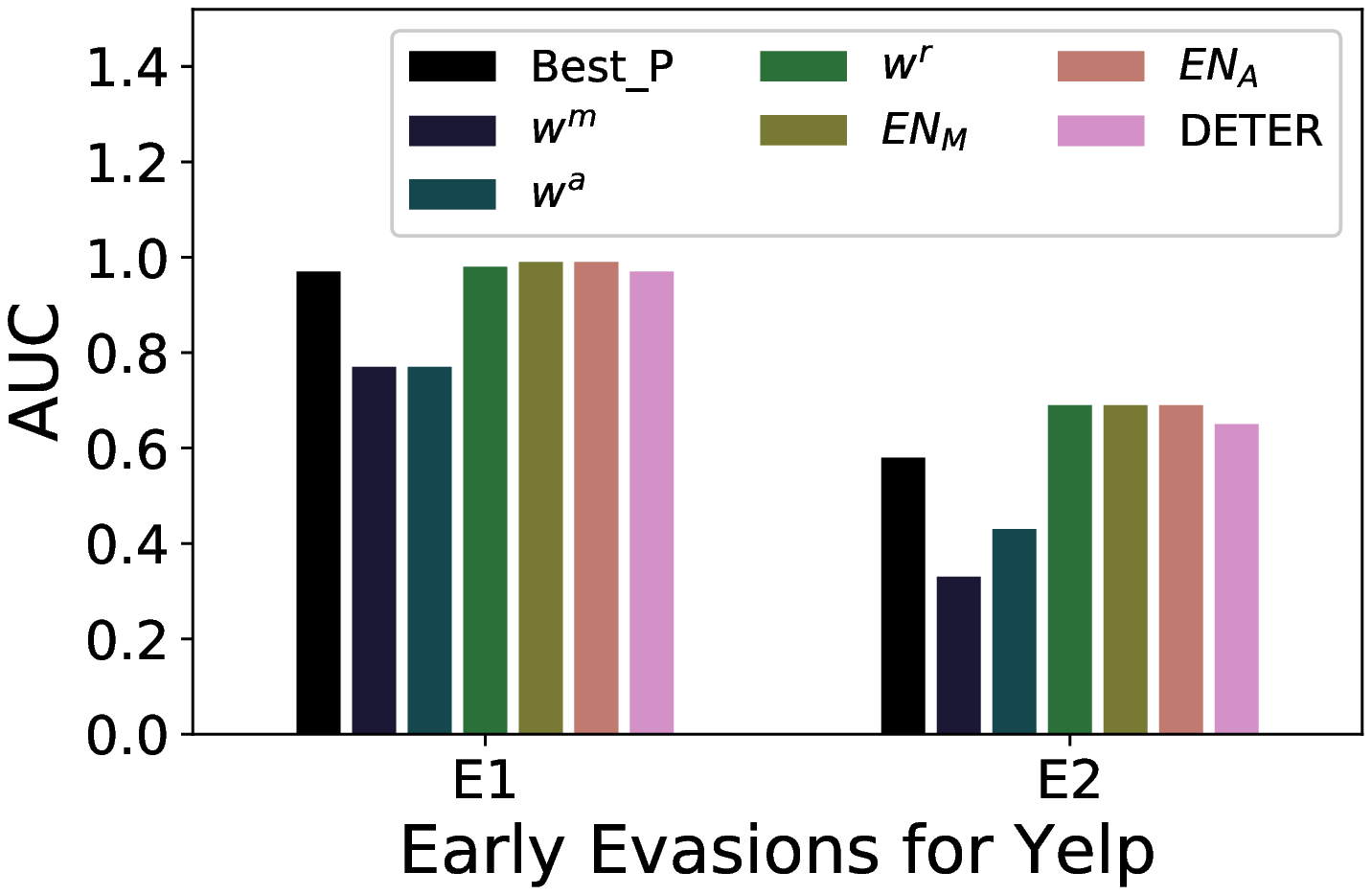}
	
}

\vspace{-.4in}
\end{table*}

The defender may assume a fixed evasion strategy that is optimal for the spammer
and then devise a detection model accordingly.
For example,
based on the above analysis,
the defender can assume that
a rational spammer will only use Evasion 4 to spam late review windows.
In reality, multiple spammers can choose different evasion strategies and a spammer can change its strategy as well.
We propose such a defense called DETER (Defense via Evasion generaTion using EmeRal)
that works well regardless of which evasion strategy is adopted by the spammers.

DETER is based on defense model re-training~\cite{liang2017, Li2016, Kantchelian2016, Goodfellow2015}.
However, the re-training relies on a evasion generator,
which is not available without EMERAL.
For a target with long review history (with more than 30 windows),
earlier windows (the first 30 windows) are used to train an EMERAL model,
which generates 9 types of evasions on later windows
(after the 30-th windows, group 1) of the targets.
For early spamming, all targets are partitioned into two subsets.
We pool all early windows (the first 30 ones) of the targets in the first subset together to train a single EMERAL model, which
generates two types of early evasions (Section~\ref{sec:early_evasion})
on all early windows of the targets in the second subset. 
Detection signals are computed for each window
where evasion are attempted.
Labels are assigned accordingly
(``spammed'' (or ``not spammed'') if EMERAL finds an evasion (or fails to generate any spam)).
For the two groups of windows with attempted evasion, respectively,
the defender pools the labeled windows
from all targets for all evasion types within the group
to train a logistic regression model (using sklearn with the default hyper-parameters) to detect windows
spammed with unknown evasion strategy during late or early review periods.

\subsection{Effectiveness of DETER for late spam detection}
Suspicious window detection AUC is used as the defender's metric~\footnote{the defender cares about both precision
and recall, while a spammer cares only about promotion effect, which is related to recall but not precision.}.
As two baselines,
the defender can train a classifier
using data obtained from each evasion type,
and during testing, detect spammed windows by pooling all classifier outputs using the MAX or AVG function
(denoted by EN\_M and EN\_A, short for ENsemble Max  and ENsemble Average, respectively).
DETER, EN\_M and EN\_A are all based on retraining and agnositic about evasion strategies,
as opposed to $\mathbf{w}^i,i=1,\dots, 9$, $\mathbf{w}^a$, $\mathbf{w}^m$ and $\mathbf{w}^r$, which
are only best for a single evasion strategy.

Two randomized evasion strategies are created to confirm that DETER works without knowing the evasion strategy.
The first one (``Rand1'')
assumes that each window is spammed with one of the 9 pure strategies with equal probability,
and the second (``Rand2'') assumes that half of the windows are spammed 
with Evasion 4, while the remaining windows
are spammed with the other strategies with equal probability.
Overall, there are 11 evasion strategies (9 pure: E1 to E9, plus 2 mixed: Rand1 and Rand2)
and 15 defense strategies (9 pure: $\mathbf{w}^{i},i=1,\dots,9$,
plus $\mathbf{w}^m$, $\mathbf{w}^a$, $\mathbf{w}^r$, EN\_M, EN\_A and DETER),
resulting in 11$\times$15 strategy profiles.
Rand1, Rand2, E5 - E9 are randomized algorithms
and we repeat each evasion and detection for 10 times,
and the means 
of the AUCs under each strategy profile are reported in Table~\ref{tab:all_defense}.
Standard deviation of AUC greater than $5e-3$ are reported as the subscripts
of the means.
Evasion strategies E1 - E4 are deterministic and only one experiment is needed.
Due to space limit, we show the best AUC of $\{\mathbf{w}^i, i=1,\dots, 9\}$ (Best\_P).

From the table, we have the following observations.
First, under strategies Rand1 and Rand2, DETER has the highest AUC than all the remaining defenses.
Among the agnostic defenses, by averaging, EN\_A is the runner-up beating EN\_M,
indicating that taking the maximum of the output is a reasonable defense but
can be over-sensitive.
Second, Best\_P is always better than $\mathbf{w}^m$, $\mathbf{w}^a$ and $\mathbf{w}^r$, and we conclude that
if the defender knows the exact evasion strategy,
it can pick a single detection signal, rather than guessing using $\mathbf{w}^r$, which is inferior to DETER.
Third, under E3, E4, E7 and E9 on the Amazon dataset,
and E5 to E9 on the Yelp dataset,
Best\_P outperforms all agnostic strategies (indicated by bold fonts with asterisks).
However, such performances are based on the unrealistic assumption
that all windows are spammed with the specific evasion strategies,
and cannot be achieved in reality.
According to Figure~\ref{fig:evasion_properties},
E5 to E9 are not effective in promoting target reputations (unprofitable for spammers)
and a spammer is less likely to select them, although DETER outperforms or is comparable to Best\_P.
The take-away is that,
by evasive spamm generation, data pooling and detection model retraining,
a defender can achieve state-of-the-art detection performance.

\begin{figure}[t]
\centering
\subfloat[Amazon\label{fig:deter_w_amazon}]{\includegraphics[width=0.23\textwidth]{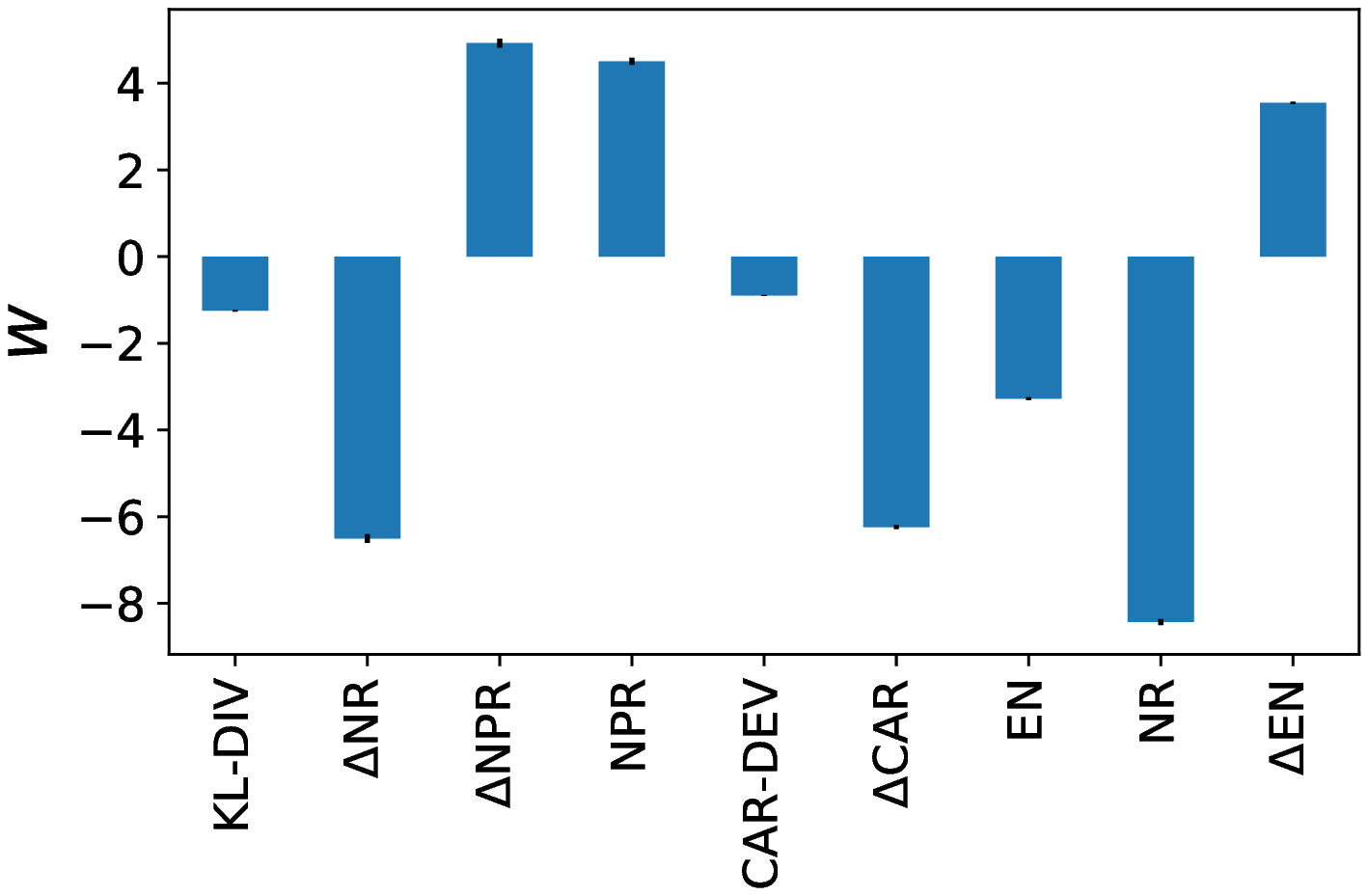}}
\subfloat[Yelp\label{fig:dter_w_yelp}]{\includegraphics[width=0.23\textwidth]{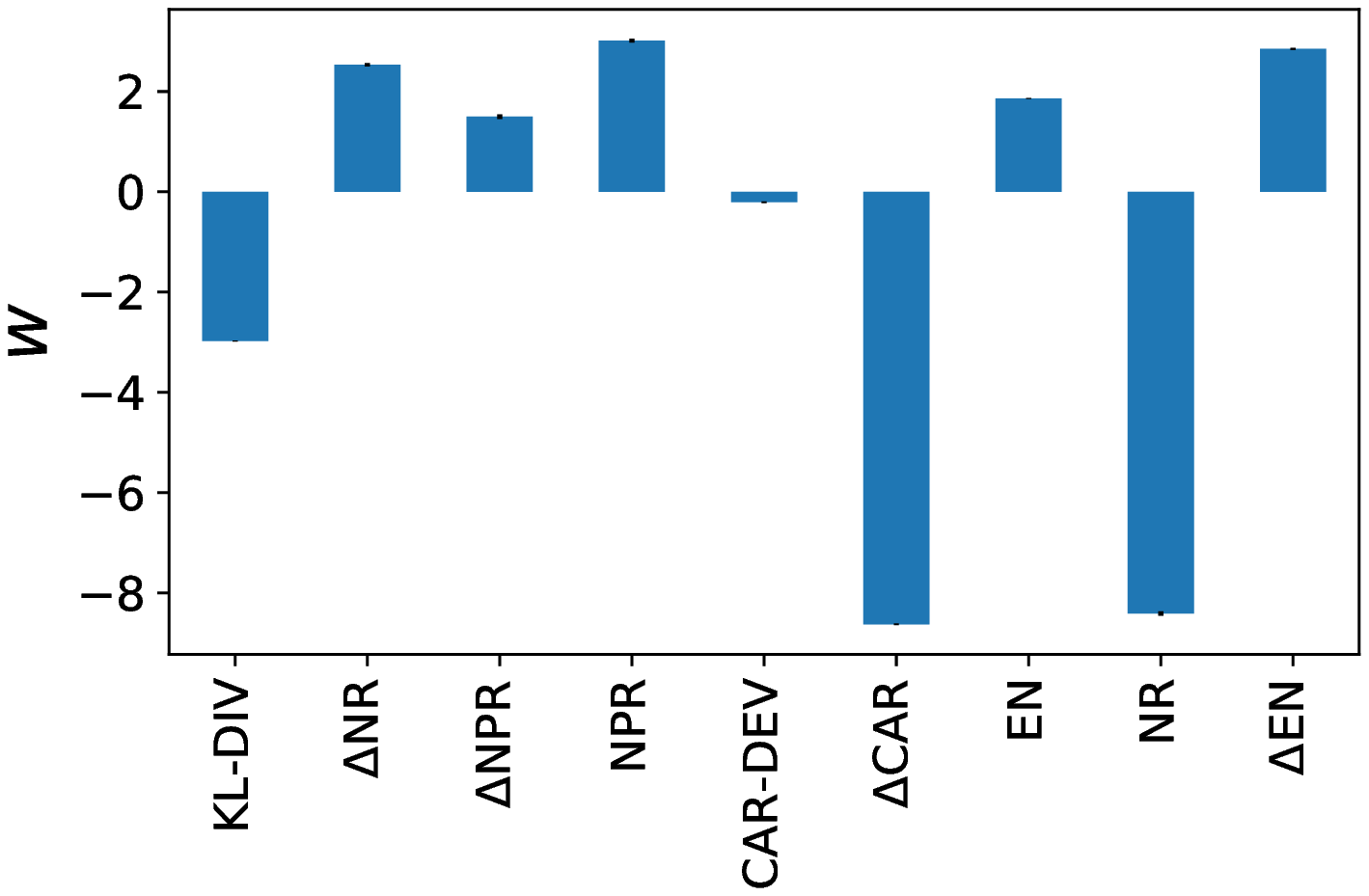}}
	\vspace{-0.5em}
\caption{\small
	Weights over detection signals learned by DETER.
}
\label{fig:deter_weights}
\vspace{-.1in}
\end{figure}

One may question the security of DETER: what if a spammer reads this paper and then implements and evades DETER?
Figure~\ref{fig:deter_weights} shows the weights learned by DETER over the 9 detection signals on two datasets.
We can see that CAR-DEV is not used much by DETER, but $\Delta$CAR and NR are always active to prevent the dominating evasion E4.
With other few medium weights watching rating distribution entropy, it would be quite difficult for a spammer
to evade this set of detection signals, while evading a larger set of signals will significantly reduce reputation promotion
(see Section~\ref{sec:interaction} especially Figure~\ref{fig:payoff_matrices}).
The strategy profile consisting of the trained DETER model and any evasion strategies is a Nash equilibrium
when the defender aims at detection AUC and spammers aim at promotion.

\subsection{Effectiveness of DETER for early spam detection}
For early spamming, the spammer can choose from two evasion strategies, E-A and E-B.
As shown in Figures~\ref{fig:early_auc_amazon} and \ref{fig:early_auc_yelp},
when dealing with evasion type E-A, the 3 adaptive detectors based on EMERAL and re-training (EN\_M, EN\_A and DETER)
have comparable or even better performance than the best pure detection strategy (Best\_P).
When dealing with E-B, these three detectors significantly outperform Best\_P.
In sum, EMERAL provides sufficient knowledge
about a wide spectrum of spams to vaccinate the defender
in the face of whatever evasion strategy.

\subsection{Effectiveness of DETER for spamming review detection}
\begin{figure}[t]
\centering
\subfloat{\includegraphics[width=0.23\textwidth]
{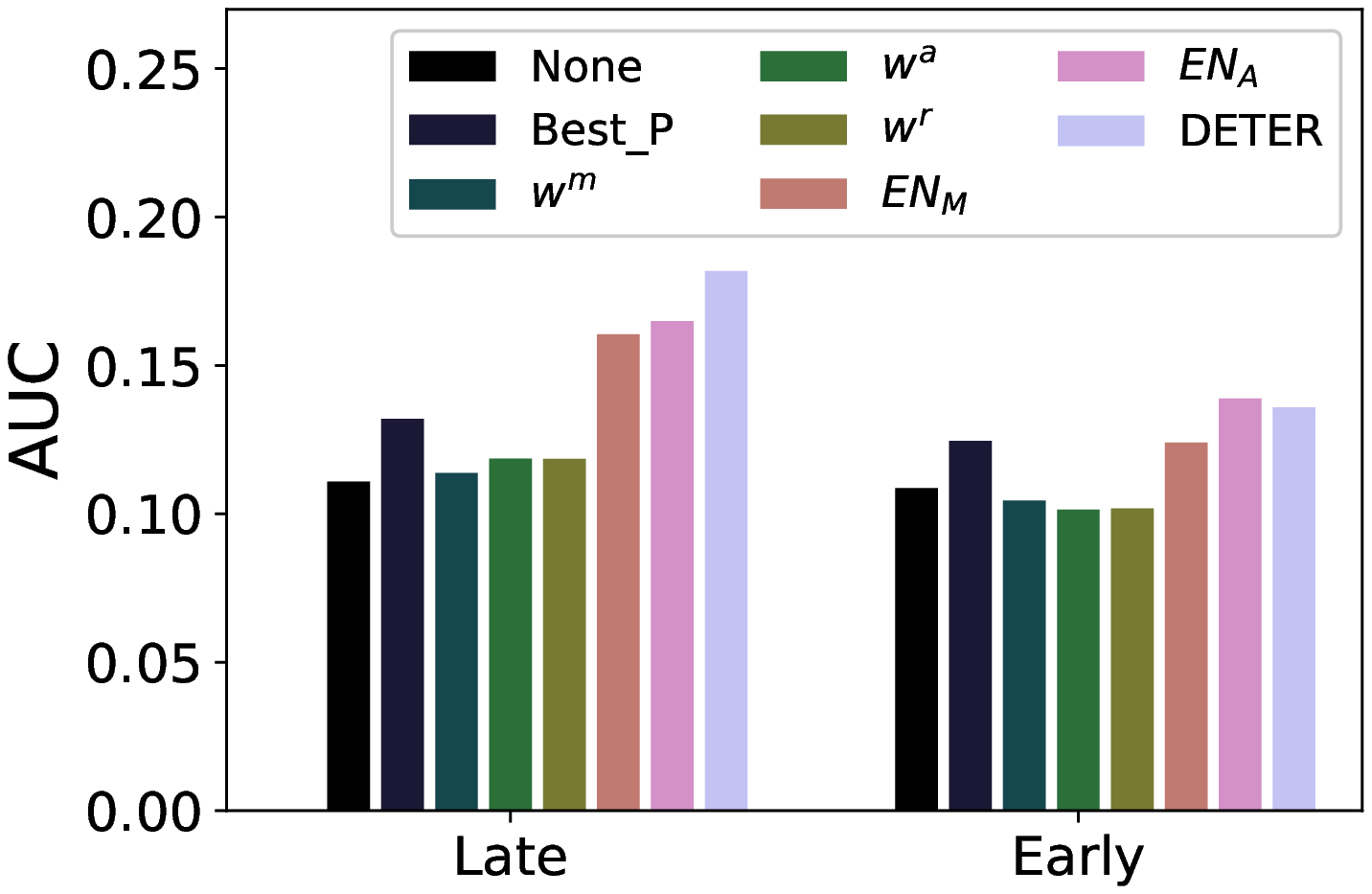}}
\subfloat{\includegraphics[width=0.23\textwidth]
{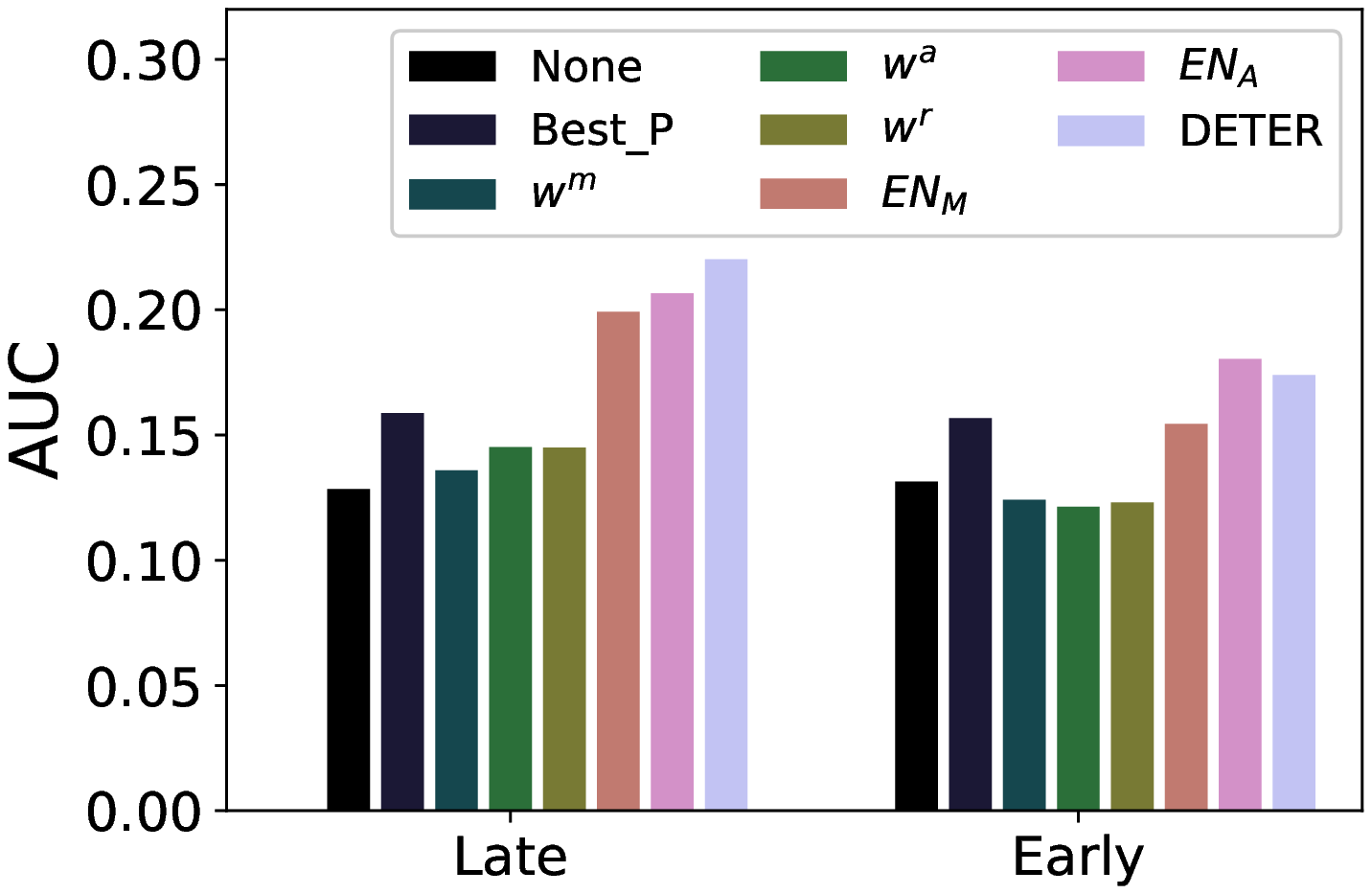}}
	\vspace{-0.5em}

\caption{Detection of spams in early and late evasion on two Yelp datasets. (Left:YelpChi, right: YelpNYC)}
\label{fig:speagle_auc}
\vspace{-.1in}
\end{figure}

We adopt the state-of-the-art spam detector, SpEagle~\cite{Rayana2015Collective},
which combine the features of the reviews, reviewers and products with the reviewer-product graph.
We show that the window suspicious scores generated by those window detectors based on re-training can help SpEagle
identify individual spamming reviews.
We run evasion E4 for late spamming and evasion E-B for early spamming on YelpChi and YelpNYC datasets~\cite{Rayana2015Collective}, respectively,
generating spams to be detected by SpEagle.
The above evasions provide the rating distributions of the spams in test windows,
and
the actual spams are posted by a random subset of the existing accounts at some randomly picked time during the test window.
To rank reviews based on their suspicious scores,
we multiply the review posteriors produced by SpEagle by the suspicious score of the window where the review sits in.
The detection AUC are shown in Figure~\ref{fig:speagle_auc}.
It is clear that those window detectors based on re-training using EMERAL (EN\_M, EN\_A and DETER) outperform the remaining ones.
In particular, DETER outperforms EN\_M and EN\_A in the late spamming cases and is comparable to EN\_A in the early spamming cases.

\section{Related Work}
\label{sec:related_work}
Opinion spams are different from
social spams~\cite{Hu14a, Grier2010, Lee2010, Yang2011},
web spams~\cite{Wu2005},
email spams~\cite{sahami98} in terms of spamming goal and detection mechanism,
and we focus on opinion spams.
Graph-based approaches leverage the relationships between reviewer accounts,
reviews and products to detect spams, suspicious accounts and dishonest businesses~\cite{wang2011review, li2014spotting, Rayana2015Collective}, even with evasive camouflages~\cite{hooi2016fraudar}
Text-based approaches identify spamming reviews based
on the contents of the reviews, using linguistic features and psychological features~\cite{ott2011finding},
topic model~\cite{sandulescu2015detecting}, semantic analysis~\cite{kim2015deep}, etc.
Behavior-based approaches~\cite{Xie2012a, Fei13, Ye2016, Santosh2016, lim2010, Jindal2007}
look for abnormal patterns in
the the volume and distribution of user ratings,
which 
are complementary to
graphs and texts based approaches.
To the best of our knowledge,
no previous work has considered generative models
for evading and securing behavior-based opinion spam detection.

Randomized defenses help
to obfuscate the details of the defender and prevent attackers from taking advantage
of any static defense strategies~\cite{Rota2017, Vorobeychik2014}.
DETER does not need randomization for privacy purpose,
since it prevents the spammers from creating campaigns that are \textit{both} evasive and effective.
If privacy is indeed a concern when using DETER,
randomization can be implemented via differential privacy~\cite{Chaudhuri2009}.
Randomized evasion is handled by DETER, demonstrated by two randomized evasion strategies.

Generating adversarial examples
is critical to secure and robust machine learning models:
if the models can see (foresee) most/all of the adversarial examples during training~\cite{Li2016, liang2017},
then during test time, most adversarial examples crafted by the attackers
can be correctly detected.
Adversarial example can be generated in either feature spaces~\cite{Lowd2005, Huang2011, Vorobeychik2014} or problem spaces~\cite{Vorobeychik2014, Xu2016, Srndic2014}.
Generation in the feature space usually admits a convex and differentiable optimization problem
whose solutions can be efficiently found as adversarial examples.
However, the generated vectors usually cannot be mapped to realistic examples in the problem space
and often tend to be over-pessimistic.
Example generation in problem spaces requires domain knowledge and usually involves non-convex and non-differentiable optimization problems.
The work here is the first step towards rigorous, efficient and realistic
adversarial spam generation in the problem space.

Game theory has been used in secure machine learning~\cite{Rota2017, Lowd2005, Bruckner2009}.
They assume that the attacker and defender know each other's objective function
and try to use game theory to arrive at a Nash equilibrium so that both parties do not seek other solutions.
We use the concept of game theory to analyze the behaviors of a rational and well-informed spammer,
instead of using game theory to find a secure defense solution.
In fact, a Nash equilibrium may be too strong an assumption in the context of spam detection,
as multiple spammers can adopt different strategies or a spammer may have no knowledge about the defender's strategy.

\section{Conclusion}
We proposed a flexible and general computational evasion model (``EMERAL'') against state-of-the-art spam detection techniques 
for both early and late stage review periods of the targets.
The spamming campaigns generated are effective in reputation manipulation
and detection evasions,
and require only public available datasets and published detection methods without knowing the exact hyper-parameter values.
EMERAL does not require differentiable models or heuristic search.
We showed that a spammer can only evade a handful of signals
but has a dominating evasion strategy representing the worst case for the defender.
We considered more realistic scenarios with mixtures of evasion strategies,
and devised DETER, an evasion-agnostic defenses based on model retraining.
Experiments showed that data pooling is the best defense,
among other ensemble methods.

\bibliographystyle{plain}
\bibliography{paper} 

\end{document}